\documentclass[lettersize,journal]{IEEEtran}
\usepackage{amsmath,amsfonts}
\usepackage{algorithmic}
\usepackage{algorithm}
\usepackage{array}
\usepackage[caption=false,font=normalsize,labelfont=sf,textfont=sf]{subfig}
\usepackage{textcomp}
\usepackage{stfloats}
\usepackage{url}
\usepackage{verbatim}
\usepackage{graphicx}
\usepackage{cite}
\usepackage{titlesec}
\titlespacing\section{0pt}{6pt}{2pt}
\titlespacing\subsection{0pt}{4pt}{2pt}

\usepackage{color}
\definecolor{citecolor}{RGB}{34,139,34}
\usepackage[pagebackref=true,breaklinks=true,colorlinks,linkcolor=black,citecolor=citecolor,bookmarks=false,urlcolor=citecolor]{hyperref}
\usepackage{makecell}
\usepackage{array}
\usepackage{booktabs}
\usepackage{multirow}

\begin{document}

\title{Vox-Fusion++: Voxel-based Neural Implicit Dense Tracking and Mapping with Multi-maps}

\author{Hongjia Zhai$^*$, Hai Li$^*$, Xingrui Yang$^*$, Gan Huang, Yuhang Ming,\\ Hujun Bao,~\IEEEmembership{Member,~IEEE} and~Guofeng Zhang,~\IEEEmembership{Member,~IEEE}

\thanks{$^*$ Equal contribution}
\thanks{Hongjia Zhai (E-mail: zhj1999@zju.edu.cn), Hai Li (E-mail: garyli@zju.edu.cn), Gan Huang (E-mail:huanggan@zju.edu.cn), Hujun Bao (E-mail: baohujun@zju.edu.cn) and  Guofeng Zhang ( E-mail: zhangguofeng@zju.edu.cn) are with the State Key Lab of CAD\&CG, Zhejiang University.}
\thanks{Xingrui Yang (E-mail: yangxingrui@cardc.cn) is with High-speed Aerodynamics Institute, CARDC, China.}
\thanks{Yuhang Ming (E-mail: yuhang.ming@hdu.edu.cn) is with the School of Computer Science, Hangzhou Dianzi University, China.}
\thanks{This work was partially supported by Key R\&D Program of Zhejiang Province (No. 2023C01039).}
}




\maketitle



\begin{abstract}
In this paper, we introduce Vox-Fusion++, a multi-maps-based robust dense tracking and mapping system that seamlessly fuses neural implicit representations with traditional volumetric fusion techniques.
Building upon the concept of implicit mapping and positioning systems, our approach extends its applicability to real-world scenarios.
Our system employs a voxel-based neural implicit surface representation, enabling efficient encoding and optimization of the scene within each voxel. 
To handle diverse environments without prior knowledge, we incorporate an octree-based structure for scene division and dynamic expansion.
To achieve real-time performance, we propose a high-performance multi-process framework. 
This ensures the system's suitability for applications with stringent time constraints.
Additionally, we adopt the idea of multi-maps to handle large-scale scenes, and leverage loop detection and hierarchical pose optimization strategies to reduce long-term pose drift and remove duplicate geometry.
Through comprehensive evaluations, we demonstrate that our method outperforms previous methods in terms of reconstruction quality and accuracy across various scenarios. 
We also show that our Vox-Fusion++ can be used in augmented reality and collaborative mapping applications. Our source code will be publicly available at \url{https://github.com/zju3dv/Vox-Fusion_Plus_Plus}
\end{abstract}

\begin{IEEEkeywords}
Dense SLAM, Implicit Representation, Voxelization, Surface Rendering.
\end{IEEEkeywords}

\section{Introduction}
\label{sec:introduction}
\IEEEPARstart{D}{ense}
simultaneous localization and mapping (SLAM) is the task of estimating a moving camera while reconstructing a dense representation of the surrounding environments. It is a key technology behind many real-world applications, such as trajectory prediction~\cite{indoor_traj}, augmented reality (AR) and virtual reality (VR)~\cite{hand-recon}. 
Dense SLAM, compared to its sparse counterpart, offers better scene completeness and more accurate camera positioning and, therefore provides better results on handling scene occlusions and collision detection.

Dense scene representation can be roughly classified into three different types, cost volumes~\cite{newcombe:2011:dtam, weerasekera:2019:gooddense} associated with keyframes, surface elements (surfels)~\cite{stuckler:2014:multisurfel, whelan:2015:efusion, wang:2019:densesurfel} or voxels~\cite{newcombe:2011:kinfu, niebner:2013:voxelhashing, kahler:2015:inftam, yang:2022:fdslam}. Since dense SLAM aims to reconstruct a complete scene geometry with densely connected topology, these systems usually need to employ a graphics processing unit (GPU) to accelerate computation to reach real-time performance~\cite{high-performance}.
While they show promising results on reconstructing room-scale scenes, their large computational requirements pose a challenge to edge computing devices and mobile robotics platforms. It is also difficult to render novel views from unseen viewpoints due to the lack of ability to complete missing geometry and texture. 

The key to better dense SLAM systems is therefore reducing memory consumption and improving the efficiency of the mapping process. 
Utilizing differentiable pipelines such as computational graphs works like CodeSLAM~\cite{bloesch:2018:codeslam} and its subsequent developments~\cite{czarnowski:2020:deepfactors, matsuki:2021:codemapping} have demonstrated the potential of neural networks in encoding depth maps with optimizable latent embeddings. 
These latent codes can be updated with multi-view constraints, striking a balance between scene quality and memory consumption. Despite their advantages, these systems have some limitations. The pre-trained networks used in these approaches often struggle to generalize well to different scene types. Moreover, obtaining a consistent global representation is challenging due to the reliance on local latent codes, which may hinder the overall accuracy and coherency of the map.
They also did not address the novel view synthesis problem.

Recent advances on neural radiance fields (NeRF)~\cite{mildenhall:2020:nerf} show potential for learning more detailed scenes with a single multi-layer perceptron (MLP). Starting from iMap~\cite{sucar:2021:imap}, there are already many studies~\cite{zhu:2021:niceslam,co-slam, vox-fusion} focusing on dense visual tracking and mapping using only neural fields. Neural fields are omnipotent function approximators that can be used to encode continuous scene properties, such as radiance, density, signed distance values, etc., which is very suitable for dense SLAM. Also leveraging the power of differentiable rendering, camera poses can be jointly optimized within the same pipeline. While existing ``NeRF-SLAM'' approaches perform well on reconstructing indoor scenes, there are two fundamental constraints that limit their use in real-world applications. Firstly, the network and embeddings have to also encode empty space which leads to low computational efficiency. Secondly, deforming the scene to incorporate loop closure correction is a difficult task. 

To address the above challenges, We propose a neural implicit RGB-D SLAM with a hybrid scene representation and multi-map construction. For individual maps, we combine an explicit sparse voxel map embedding space~\cite{Vox-Surf,liu:2020:nsvf} with a neural implicit network. 
We then subdivide different parts of the map into multi-maps and use appearance and geometry loop constraints to join them together. 
Our approach differs from previous methods in the following aspects: (1) Unlike prior works that focus on reconstructing scenes with known bounds, we dynamically expand our sparse voxel map on the fly, making our approach more practical and applicable to real-world scans. The use of sparse voxel grids also significantly reduced memory consumption; (2) We do not use pre-trained networks. Our system learns a neural field from scratch, which ensures the reconstructed surface is not influenced by bias in the training data; (3) We incorporate loop detection from appearance and geometric features. We propose a multi-map based hierarchical loop optimization strategy that contributes to more accurate geometry and appearance for larger scenes; (4) The combination of dynamic voxel expansion and multi-map based loop correction enables us to reconstruct large indoor scenes and collaborative mapping, which is not possible for previous works. 
In contrast to the preliminary conference version~\cite{vox-fusion}, we introduce several improvements for mapping large indoor scenes in~\autoref{sec:multi-map}, which includes incremental mapping with multi-maps, appearance-based and geometry-based loop detection, and hierarchical pose optimization. Additionally, we update some training parameters for the Replica and ScanNet datasets. This update further enhances the performance of our system on these specific datasets, resulting in more accurate pose estimation and scene reconstructions.

In summary, our contributions are:

\begin{enumerate}
    \item We propose Vox-Fusion++, a novel fusion system for real-time implicit tracking and mapping. Vox-Fusion++ combines explicit voxel embeddings, indexed by a dynamic octree, and a neural implicit network, enabling scalable implicit scene reconstruction with detailed geometry and color.
    \item Our system directly renders signed distance volumes, resulting in improved tracking accuracy and reconstruction quality compared to current state-of-the-art systems, without introducing additional computation cost.
    \item To address the challenge of reconstructing large scenes, we propose an incremental approach using multi-maps. Loop detection and hierarchical pose optimization are performed between different maps, effectively reducing long-term drift and eliminating duplicate geometry.
    \item We conduct extensive experiments on both synthetic and real-world scenes to demonstrate the effectiveness of our proposed method in generating high-quality 3D reconstructions. The property of our work benefits various augmented reality and collaborative mapping applications.
\end{enumerate}


\section{Related Work}
\label{sec:related_work}

\noindent\textbf{Traditional Dense Visual SLAM:} DTAM~\cite{newcombe:2011:dtam} proposed the first dense SLAM system to track a handheld camera while reconstructing a dense cost volume representation by considering multi-view stereo constraints.
However, their method was limited to small workshop-like spaces.
Leveraging RGB-D cameras, KinectFusion~\cite{newcombe:2011:kinfu} introduced a novel reconstruction pipeline, extending dense SLAM to room-scale scenes. Benefiting from accurate depth acquisition from commodity depth sensors and the parallel processing power of modern GPUs. They employed iterative closest point (ICP) to track input depth maps and progressively updated a voxel grid with aligned depth maps. A frame-to-model tracking method was also proposed, effectively reducing short-term drifts, especially in circular camera motion scenarios. Subsequent research built upon this foundation and improved the systems by introducing different 3D structures~\cite{whelan:2015:efusion}, exploring space subdivision~\cite{roth:2012:mkinfu, niebner:2013:voxelhashing}, and performing global map optimization~\cite{kerl:2013:dvo, kahler:2016:inftam2, schops:2019:badslam}.
Another notable research direction involves combining features and dense maps~\cite{li:2015:rgbdreloc, dai:2017:bundlefusion, yang:2022:fdslam,Scene_complete,mo-slam,sf-depth}, significantly enhancing the robustness of iterative methods. 
Traditional dense SLAM systems produce impressive scene reconstructions with real-time speed, but they have two main drawbacks: large memory footprint and cannot be used to render novel views.

\noindent\textbf{Learning-based Dense Visual SLAM:} Utilizing learned geometric priors, DI-Fusion~\cite{huang:2021:difusion} encodes local points into a low-dimensional latent space, which can then be decoded to generate signed distance function (SDF) values with a single MLP. 
However, due to measurement noise and pre-trained data bias, the learned geometric prior may be inaccurate in representing complex shapes, leading to subpar surface reconstruction quality.
CodeSLAM~\cite{bloesch:2018:codeslam} adopts a U-shaped encoder-decoder structure to embed depth maps as low-dimensional codes. 
These codes, combined with a pre-trained neural decoder, allow for joint optimization of keyframes and camera poses. 
Nonetheless, similar to other learning-based methods, their approach lacks robustness in handling scene variations.
Some other works~\cite{teed2022deep,teed:2021:droid} exploit deep learning techniques for pixel-level dense matching or patch-level sparse matching. These systems operate through alternations between motion updates and bundle adjustments. 
However, especially in the case of~\cite{teed:2021:droid}, the estimation of dense motion fields between selected frames results in high computational costs and significant memory resource requirements. 
They also produce point clouds as the global map, which is only suitable for certain applications.

\begin{figure*}
\begin{center}
\includegraphics[width=0.9\textwidth]{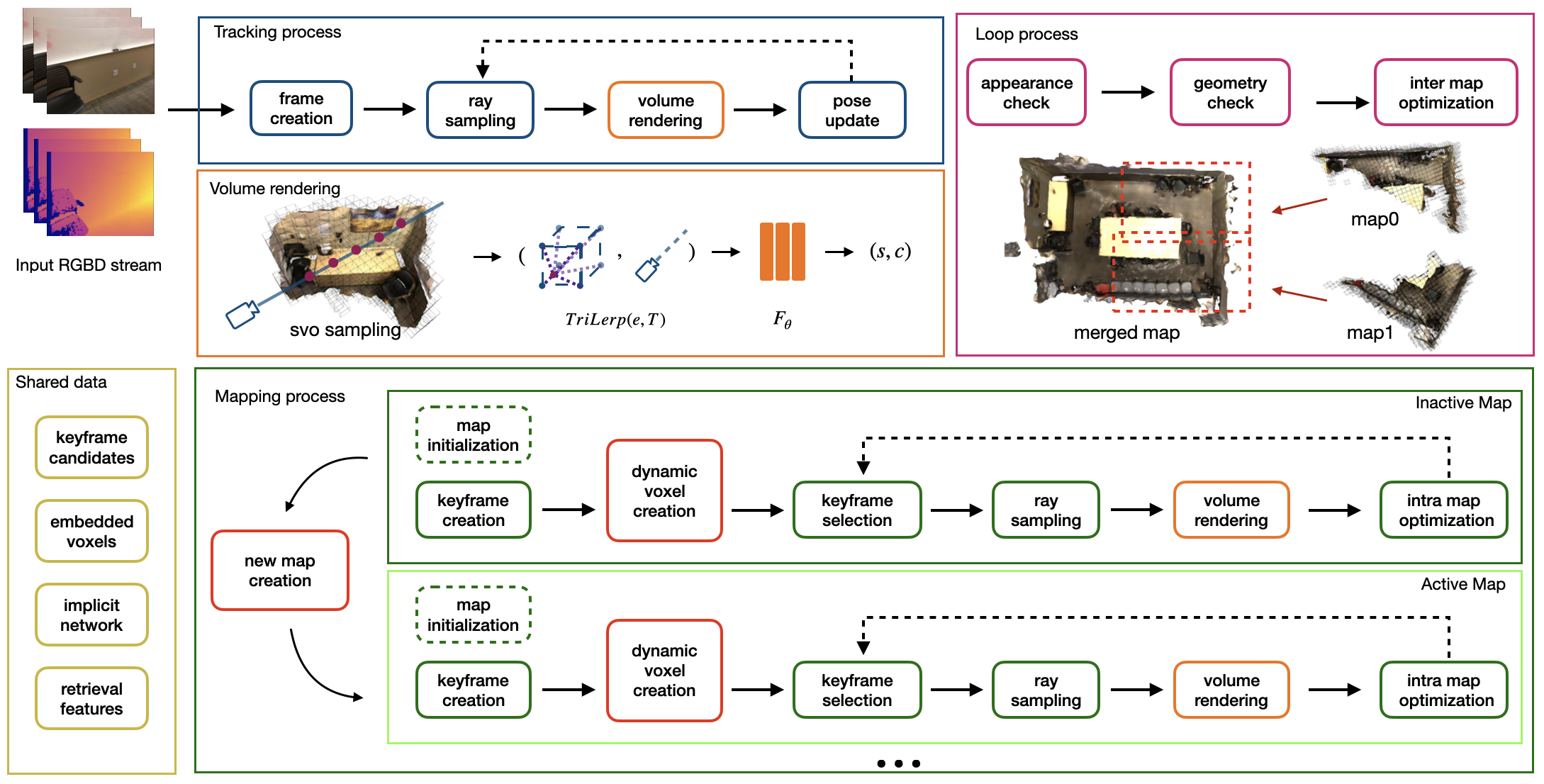}
\end{center}
  \caption{Overview of our SLAM system. The whole system consists of four parts: 1) Tracking Process: Taking RGB-D frames as input and optimizing camera poses through differentiable rendering, 2) Volume Renderer: This module encodes the scene in a MLP and voxel feature embedding, producing rendered color and Signed Distance Function (SDF) values for each point, 3) Mapping Process: Reconstructing the geometry of the scene via volume rendering and perform incremental mapping with multi-maps for large scenes, 4) Loop Process: Performing loop detection and hierarchical pose optimization between different maps to reduce the pose drift.}
\label{fig:overview}
\end{figure*}
\noindent\textbf{Neural Implicit Dense SLAM:} Recently, there have been successful approaches to representing scenes using neural implicit radiance fields~\cite{sucar:2021:imap,co-slam,zhu:2021:niceslam,vox-fusion,nicer-slam,nerf-slam,dim-slam,Orbeez-slam,eslam}.
iMAP~\cite{sucar:2021:imap} stands out as the first neural implicit SLAM approach that formulates the dense SLAM problem as a continuous learning paradigm. To improve optimization time, they employ heuristic sampling strategies and keyframe selection based on information gain, achieving a good balance between compactness and accuracy.
Addressing scalability concerns, NICE-SLAM~\cite{zhu:2021:niceslam} subdivides the world coordinate system into uniform grids. However, the use of a multi-level dense feature grid to represent scenes may result in redundant voxels in empty space. It also requires knowing the scene bound (\textit{a prior}), making it difficult to adapt to robotics applications. To address this issue, Vox-Fusion~\cite{vox-fusion} employs dynamically expanding voxel grids indexed by an octree structure, which works very well in mapping unknown scenes.
Depending on the use cases, neural implicit SLAM usually can go in two directions. One is the maximize rendering quality. Works such as~\cite{Orbeez-slam, nerf-slam} embedding neural rendering framework inside traditional SLAM pipelines~\cite{orb-slam2,orb-slam3} to utilize the full potential of existing systems. They are more time efficient and obtain novel view renderings with good quality. However, they pay little attention to geometry, making their system not suited for dense SLAM applications. Other systems~\cite{nicer-slam,dim-slam} work on RGB images. Taking advantage of existing depth estimation pipelines and multi-view consistency, they were able to get plausible reconstruction from monocular cues, albeit at the expense of longer reconstruction time. Utilizing the inductive bias of frequency encoding, Co-SLAM~\cite{co-slam} uses a hybrid of frequency and hash encoding to obtain improved geometry smoothness. 
Despite showing promising results on small-scene reconstruction, the above methods do not take loop detection into consideration, which is a central requirement for mapping larger scenes.
Our work addresses this issue by further subdividing the scene into multi-maps, and then fusing them to obtain a complete map. 
This allows us to incorporate explicit loop closing cues.

\noindent\textbf{Neural Implicit Representations:} NeRF~\cite{mildenhall:2020:nerf} introduces a method to render scenes as volumes with density, which is beneficial for representing transparent objects. 
However, their focus is on rendering photo-realistic images, and a good surface reconstruction is not always guaranteed. 
For most AR tasks, it is crucial to accurately identify the surface.
To address the surface reconstruction problem, several new methods~\cite{go-surf,manhattan-sdf,oechsle:2021:unisurf} propose implicit surface representations or use depth as supervision to achieve better surface reconstruction. These methods utilize iterative root-finding~\cite{oechsle:2021:unisurf}, weight rendered color with associated SDF values~\cite{wang:2021:neus}, or encourage the network to learn more surface details within a predefined truncation distance~\cite{azinovic:2022:neuralrgbd}. In our approach, we adopt the rendering method from~\cite{azinovic:2022:neuralrgbd}, but instead of regressing absolute coordinates, we work with interpolated voxel embeddings.
Using a single network often comes with limitations in capacity and scalability, especially when dealing with larger scenes, as it requires a significant increase in the number of learnable parameters. To address this issue, some works utilize hybrid representations that combine voxel~\cite{vox-fusion,Vox-Surf,yu:2022:plenoxels,liu:2020:nsvf}, octree~\cite{yu:2021:plenoctrees,takikawa:2021:nglod,h2mapping}, dense grid~\cite{go-surf,zhu:2021:niceslam, h2mapping} with coordinate encoding. 
For instance, NSVF~\cite{liu:2020:nsvf} embeds local information in a separate voxel grid of features, generating comparable or even better results with fewer parameters. Plenoxels~\cite{yu:2022:plenoxels} employs spherical harmonic functions as voxel embeddings, eliminating the need of neural network. 
Furthermore, an explicit feature grid~\cite{yu:2022:plenoxels,liu:2020:nsvf,imtooth} offers faster rendering speed, as the implicit network can be much smaller compared to the original NeRF network. This uniform grid design can be found in other neural reconstruction methods like NGLOD~\cite{takikawa:2021:nglod}, which utilizes a hierarchical data structure by concatenating features from each level to achieve scene representation with different levels of detail. 
\section{System overview}
\label{sec:overview}
The system overview is shown in~\autoref{fig:overview}. 
Our system takes continuous RGB-D frames as input $\mathcal{F}_i\doteq\{I_i,D_i\}$, comprising RGB images $I\in\mathbb{R}^3$ and depth maps $D\in\mathbb{R}$.
We use the calibrated depth camera with the intrinsic matrix $K\in \mathbb{R}^{3\times3}$.
During map initialization, we construct the global map by running a few mapping iterations on the first frame. 
In the tracking process, we estimate the current camera pose (6-DoF) $T\in SE(3)$ \textit{w.r.t.} the fixed implicit scene network $F_{\theta}$ using differentiable volume rendering. 
Then, each tracked frame is sent to the mapping process for constructing the global map. 
In the mapping process, we back-project and transform 3D point clouds derived from the depth maps to create the new voxel-based scene based on the estimated camera poses from the tracking process.
These newly constructed scenes are fused into the global map, followed by intra-map joint optimization.
To manage optimization complexity, we maintain a limited number of keyframes selected based on the ratio of observed voxels. 
To ensure long-term map consistency, we perform continuous optimization on a fixed window of keyframes.
For scenes that surpass a certain size, we incrementally create new maps for unexplored areas. To mitigate pose drift, loop detection, and hierarchical pose optimization between different maps are performed.
By elaborating on these components, our system achieves efficient and accurate mapping of large indoor scenes.
These individual components will be explained in detail in the following sections. 


\section{Implicit SLAM using Sparse Voxels}
\label{sec:method}
In this section, we detail the process of neural implicit tracking and mapping using sparse voxels. We first explain our volume rendering pipeline, then we analyze our tracking and mapping processes. 
Lastly, we explain our dynamic voxel managing strategy.

\subsection{Volume Renderer}
\label{sec:render}
\noindent\textbf{Voxel-based Sampling:} Following the formula of~\cite{Vox-Surf}, we construct the main structure of our map as sparsely distributed latent embeddings. These embeddings can be indexed by an octree structure using tri-linear interpolation. The interpolated voxel embeddings can then be used to decode the stored information, using an implicit SDF decoder $F_{\theta}$ learned on-the-fly. This structure has many nice properties, \textit{e.g.}, the geometry and texture are smoother around voxel boundaries, and the network can have fewer layers due to the information is partially stored in the map itself.

To avoid sampling points in the empty space, we adopt the sparse sampling strategy. For each sampled camera ray, we first perform a ray-voxel intersection test to see if they hit any voxels. If they do, we perform a ray marching using fixed step lengths to sample points along the hit voxels. 
This gives us nicely distributed point samples inside existing voxel grids. We also enforce a limit $M_h$ on the number of voxels a single ray can see to bound computation. 

However, the above method did not take scene distance into consideration. This strategy may not be ideal for complex indoor scenes, where the number of observable voxels varies a lot and no single preset value can fit all of them. We also want to minimize the impact of distant voxels where scene elements are sparse. For example, a table $20$ meters away is not likely to contribute to the rendering of a chair in front the camera. Instead of heuristically specified limits~\cite{liu:2020:nsvf, Vox-Surf}, we dynamically adjust $M_h$ based on a specified maximum sampling distance $D_{max}$ with the following procedure: We first use a large enough $M_h$ value to gather voxels, this value is pre-determined to avoid any memory overflow. We then sort the sampled points from near to far and mask out any that are too far away. This method effectively combines distance-based and number-based sampling strategies to obtain the best of both worlds.

\vspace{1mm}
\noindent\textbf{Implicit Surface Rendering:} NeRF can encode different properties such as occupancy, SDF, \textit{etc}. Previous works mostly choose occupancy since it can be trivially converted to density to fit into the rendering pipeline. However, occupancy has several drawbacks as well, the biggest problem being its inability to model thin and tiny objects. In our work, we directly regress SDF values. 
Not only because it can be used to represent thin structures, it is also a versatile format that supports other computer graphics tasks such as ray tracing. We will show our choice of SDF is better at reconstructing room-scale scenes as we learn more details in the experiment section.

The key distinction in our method is the use of voxel embeddings rather than 3D coordinates, which sets us apart from many previous approaches. 
For rendering color and depth from the sparsely sampled points in the previous section, we adopt the volume rendering method proposed in~\cite{azinovic:2022:neuralrgbd}. We made modifications so it can be applied to feature embeddings rather than global coordinates. More specifically, we do not use frequency encoding as NeRF does, and we use fewer parameters. For $N$ sampled points, we use the following rendering function to obtain the color $\textbf{C}$ and depth $D$ for each ray:
\begin{gather}\label{eq:render}
    (\mathbf{c}_i,s_i)=F_{\theta}(\text{TriLerp}(T_i\mathbf{p}_i, \mathbf{e})), \\
    w_i = \sigma(\frac{s_i}{tr})\cdot \sigma(-\frac{s_i}{tr}), \\ 
    \mathbf{C} = \frac{1}{\sum^{N-1}_{j=0}w_j}\sum^{N-1}_{i=0}w_j\cdot\mathbf{c}_j, \\
    D = \frac{1}{\sum^{N-1}_{j=0}w_i}\sum^{N-1}_{j=0}w_j\cdot d_j, 
\end{gather}
where $T_i$ represent the pose of the current frame, $\text{TriLerp}(\cdot, \cdot)$ is the trilinear interpolation function, $F_{\theta}$ is the implicit network with trainable parameters $\theta$. $\mathbf{c}_j$ is the predicted color for each 3D point from the network, by trilinearly interpolating voxel embeddings $\mathbf{e}$. Likewise, $s_i$ is the predicted SDF value and $d_j$ the $j$-th depth sample along the ray. $\sigma(\cdot)$ is the sigmoid function and $tr$ is a pre-defined truncation distance. The depth map is similarly rendered from the map by weighting sampled distance instead of colors. 

\vspace{1mm}
\noindent\textbf{Optimization Objectives:} To supervise the network, we employ four different loss functions: RGB loss, depth loss, free-space loss, and SDF loss, which are calculated based on the sampled points $P$. The RGB and depth losses are computed as the absolute differences between the rendered images and the ground-truth images:
\begin{equation}
\begin{split}
    \mathcal{L}_{rgb} = \frac{1}{\vert P\vert}\sum_{i=0}^{\vert P\vert}\| \textbf{C}_i - \textbf{C}^{gt}_i\| ,\\ \mathcal{L}_{depth} = \frac{1}{|P|}\sum_{i=0}^{\vert P\vert}\| D_i - D^{gt}_i\| ,
\end{split}
\end{equation}
where $D_i,C_i$ are the rendered depth and color of the $i$-th pixel in a batch, respectively. $D_i^{gt},C_i^{gt}$ are the corresponding ground truth values. The free-space loss works with a truncation distance $tr$ within which the surface is defined. The MLP is forced to learn a truncation value $tr$ for any points lie within the camera center and the positive truncation region of the surface:
\begin{equation}
    \mathcal{L}_{fs} = \frac{1}{\vert P\vert }\sum_{p\in P} \frac{1}{S_p^{fs}} \sum_{s \in S_p^{fs}}(D_s - tr)^2.
\end{equation}
Finally, we apply SDF loss to force the MLP to learn accurate surface representations within the surface truncation area:
\begin{equation}
    \mathcal{L}_{sdf} = \frac{1}{\vert P\vert}\sum_{p\in P} \frac{1}{S_p^{tr}} \sum_{s \in S_p^{tr}}(D_s - {D}_s^{gt})^2.
\end{equation}
In contrast to methods like~\cite{rematas:2022:urf}, which require the network to learn a negative truncation value $-tr$ for points behind the truncation region, we adopt a simpler approach by masking out these points during rendering. This avoids the challenge of solving surface intersection ambiguities~\cite{wang:2021:neus} as proposed in~\cite{azinovic:2022:neuralrgbd}. This straightforward formulation allows us to achieve accurate surface reconstructions with significantly faster processing speeds.

\subsection{Tracking Process}
\label{sec:tracking}
During the tracking process, we keep our voxel embeddings and the parameters of the implicit network fixed. In other words, we only optimize the 6-DoF pose $T\in SE(3)$ for the current camera frame. Similar to previous methods, where pose estimates are iteratively updated by solving an incremental update, in each update step, we measure the pose update in the tangent space of $SE(3)$, represented as the Lie algebra $\xi \in \mathfrak{se}(3)$.
To simplify the process, we assume a zero motion model where the new frame is sufficiently close to the last tracked frame. Consequently, we initialize the pose of the new frame to be identical to that of the last tracked frame. For each frame, we sample a sparse set of $N_t$ pixels from the input images for tracking.

\begin{equation}
    \min_{\hat{\xi}}  \mathcal{L} (\sum_{i}^{N_t}F_{\theta}(\text{TriLerp}(\hat{\xi}_iT\mathbf{p}_i, \mathbf{e})), C^{gt}, D^{gt})
\end{equation}
where $\hat{\xi}$ is the pose update of current frame, and $\mathcal{L}$ is the optimization objectives $\{\mathcal{L}_{rgb}, \mathcal{L}_{depth}, \mathcal{L}_{fs}, \mathcal{L}_{sdf}\}$. 

During the tracking process, we follow the procedure described in~\autoref{sec:render} to sample candidate points and perform volume rendering. 
The frame pose is updated in each iteration via back-propagation. 
Similar to~\cite{sucar:2021:imap}, we maintain a copy of our SDF decoder and voxel embeddings for the tracking process. 
This map copy is directly obtained from the mapping process and updated each time when a new frame has been fused into the map.

\subsection{Mapping Process}
\label{sec:mapping}
In the mapping process, we dynamically create new maps as the explored area gradually increases and maintain multi-maps for reconstruction and optimization. Here, we show the mapping process for a single map.

\vspace{1mm}
\noindent\textbf{Key-frame Selection:} In the context of online continuous learning, keyframe selection plays a vital role in maintaining long-term map consistency and preventing catastrophic forgetting~\cite{sucar:2021:imap}. Unlike previous methods that rely on heuristically chosen metrics~\cite{sucar:2021:imap} or fixed intervals~\cite{zhu:2021:niceslam} to insert keyframes, our explicit voxel structure enables a more dynamic approach. We perform an intersection test to determine when to insert keyframes, making use of the current frame's impact on the existing map. Specifically, after successfully tracking a new frame, we assess how many voxels $N_c$ would be allocated if this frame were chosen as a new keyframe. We then compute the ratio $p_{kf}=N_c/N_o$, where $N_o$ represents the number of currently observed voxels. If the ratio exceeds a predefined threshold, we insert the new frame as a keyframe.

While this straightforward strategy is effective for exploratory movements due to continuous voxel allocation, it may face challenges with loopy camera motions, especially those involving long-term loops. Such situations can lead to an inability to allocate new keyframes, resulting in missing parts of the model or insufficient multi-view constraints. To address this issue, we introduce a maximum interval between adjacent frames for keyframe insertion. If a new keyframe has not been added for the past $N$ frames, we create a new one to ensure consistent scene mapping. This keyframe selection approach is both simple and robust, ensuring a coherent and complete scene map in various scenarios.

\vspace{1mm}
\noindent\textbf{Bundle Adjustment:} In our mapping subroutine, we integrate the tracked RGB-D frames into the existing scene map by jointly optimizing the scene geometry and camera poses. To address the challenge of network forgetting in online incremental learning~\cite{sucar:2021:imap}, we employ a similar method to perform joint optimization of the scene network and feature embeddings. For each frame, we randomly select a subset of $N_{kf}$ keyframes, which includes the recently tracked frame. These keyframes form an optimization window, akin to the sliding window approach used in traditional SLAM systems~\cite{strasdat:2011:doublewindow}.

In the optimization window, we employ a similar process as in the tracking phase. For each frame, we randomly sample a set of $N_m$ rays. These rays are transformed into the world coordinate system using the estimated frame poses. Next, we sample points within our sparse voxels and render a set of pixels from these sample points. We then calculate the related loss functions, as described in~\autoref{sec:render}, to guide the joint optimization process. This iterative optimization procedure helps to refine the scene geometry and camera poses, ensuring accurate and consistent mapping of the scene during the mapping process.

\subsection{Dynamic Voxel Management}
\label{sec:allocate}
\noindent\textbf{Voxel Allocation:} We employ an off-the-shelf octree structure~\cite{vespa:2018:supereight} to efficiently manage sparse voxel maps, which enable us to create, delete and search voxels on-the-fly. Similarly one can use hash tables~\cite{niebner:2013:voxelhashing} to achieve the same goal, but we choose octree over hash table for its simplicity and the ability to be easily shared among different threads. Instead of storing SDF directly, all leaf nodes in the octree store indices of the neural embeddings for its eight corners, allowing us to efficiently index embeddings from a separate list.

When exploring a new area, we dynamically allocate voxels whenever new observations are made. More specifically, at the beginning of the process, the leaf nodes corresponding to the unobserved scene areas are set to empty. When a new frame is successfully tracked, its associated depth map is back-projected into 3D points, which are then transformed by the estimated camera pose. For any point that does not fall into an existing voxel, we allocate a new one. This approach ensures that we do not need to specify scene bound for our system to work, we gradually expand the map as we explore unknown regions, which is suitable for robotics platforms.

\vspace{1mm}
\noindent\textbf{Morton Encoding:} Depending on input resolution, we may need to process tens of thousands of points in a very short time frame. To accelerate this process, we choose to encode voxel coordinates using Morton codes. Morton codes are formed by interleaving the bits from each coordinate into a single unique number, providing an efficient and compact representation. Given the 3D coordinate $(x, y, z)$ of a voxel, we can rapidly determine its position in the octree by traversing through its Morton code. Moreover, through a decoding operation, we can easily recover the encoded coordinates. Additionally, the neighboring voxels can be identified by shifting the appropriate bits of the Morton code, which proves advantageous for swiftly locating shared embedding vectors among neighboring voxels. This choice of encoding significantly contributes to the efficiency and effectiveness of our system, particularly in terms of voxel allocation and retrieval. For a more detailed explanation, we refer interested readers to~\cite{vespa:2018:supereight}.

\section{Loop Closure based Multi-maps}
\label{sec:multi-map}
As stated in the previous sections, there are several challenges in large scene reconstruction for neural implicit SLAM.
Among them, the forgetting problem of the network model and the long-term drift of the pose estimation are the two most crucial.
To address the above-mentioned issues, we propose to incrementally reconstruct large-scale scenes with multi-maps based on our sparse voxel representation.

\subsection{Incremental Mapping with Multi-Maps}
In our multi-map system, we model the states as set of maps $\mathcal{M}\doteq\{\mathcal{E},\mathcal{K},\theta\}$, where $\mathcal{E}$ is the set of voxel embeddings, $\mathcal{K}\doteq\{I,D,T\}$ is the keyframe collection within the same map and $\theta$ is the network parameters associated with the map. We only create a new map when the reconstructed region exceeds the space $\Omega_{\mathcal{M}}$ managed by the current scene octree, or camera tracking failure is encountered.

This construction has several distinct benefits. 
Firstly, it allows us to bound computation. Our sampling strategy has a computational complexity proportionate to the number of voxels in the map (bounded by sampling distance), the same problem inherited from traditional dense voxel SLAM~\cite{niebner:2013:voxelhashing}. By specifying a region threshold, we can maintain a constant map size and, therefore more efficient on reconstructing large scenes. Secondly, we handle tracking failure and loop closure in the same pipeline. Tracking failure will always lead to a new map, which will hopefully be loop closed with the other sections of the map. 
Thirdly, the forgetting problem is mitigated by using multiple small MLPs in each map. When we create a new map, we leave the parameters of the previous map unchanged during the subsequent mapping process. Therefore, we do not have to worry about forgetting in different maps, which leads to the constant size of the key-frame selection window. Finally, our system allows us to perform loop detection and closing. 
Drifting is a long-standing problem in SLAM where small errors in each estimation step accumulate. 
Such errors, if not corrected, will lead to catastrophic failures. Leveraging inter-map and intra-map pose optimization, we model the loop closing problem as a multi-map optimization framework and efficiently reduce drifts between keyframes and maps, which has not been possible for previous works.

\begin{figure}
    \centering
    \includegraphics[width=0.9\linewidth]{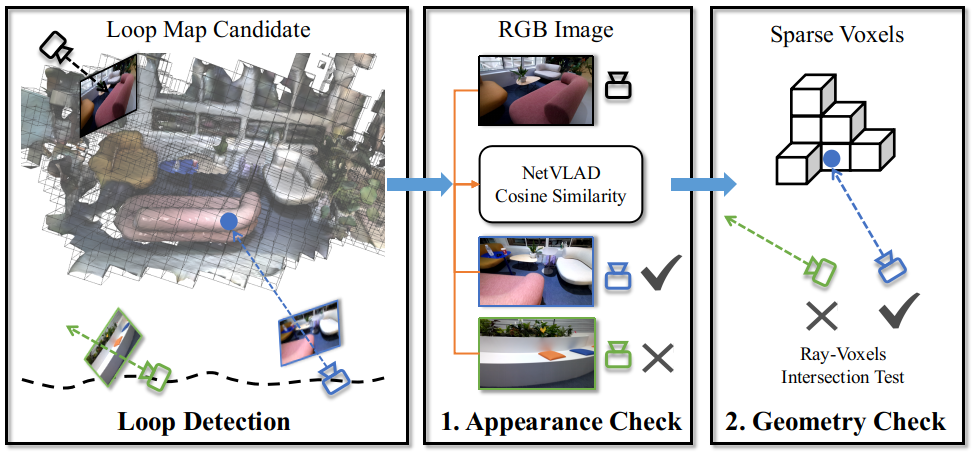}
    \caption{Illustration of our loop detection. 1) Appearance check. We first calculate the similarity between the current frame and keyframes inside the loop map candidate. 2) Geometry check. We then perform the intersection test between rays from current frame and sparse voxels of loop map.}
    \label{fig:loop_detection}
\end{figure}
\subsection{Robust Loop Detection}
Indoor scenes often contain similar and repetitive features (such as tables, chairs, \textit{etc}.), and they can cause significant trouble to visual place recognition methods~\cite{netvlad,patch-netvlad,hloc}. 
Therefore we propose to perform geometric verification after visual feature matching for robust loop detection. 
The complete process is illustrated in~\autoref{fig:loop_detection}. 

\vspace{1mm}
\noindent\textbf{Appearance Check:} 
To detect loop closure candidates, we first use the pre-trained NetVLAD~\cite{netvlad} model to extract global feature vectors for each keyframe $\{f^{kf}_i=\texttt{NetVLAD}(I_i^{kf}) \in \mathbf{R}^{512}\}$. 
The extracted features are stored in a keyframe database, along with their associated sub-map IDs. 
Then for every new frame, we extract the same type of feature and match them with every other key-frame feature using nearest neighbor searching, by calculating the pairwise similarity between each keyframe and the current frame feature vector. Here we use the cosine similarity between feature vectors as the similarity measurement. We determine the most similar keyframes whose similarity score is the greatest. To filter out outliers, we also apply a threshold check. A frame can be treated as loop candidate only if its similarity score is greater than a predefined threshold. Note that we do not perform pair-wise pose estimation at this stage, the loop frame is verified using our geometric verification method and optimized using our proposed hierarchical pose optimization scheme. 

\vspace{1mm}
\noindent\textbf{Geometry Verification:} Owing to the explicit voxel structure, our representation allows us to quickly verify if a keyframe can be trusted as correct loop constraints. 
More specifically, for each loop candidate frame obtained from the appearance check, we perform an intersection test between the rays sampled from the frame with the sparse voxels of the map. We determine the frame passes the geometry check if there exists voxel intersections. Due to the relatively small region of each multi-map, this method is accurate enough to filter out invalid frames to prevent false positives.

\begin{figure*}
\centering 
\includegraphics[width=0.9\textwidth]{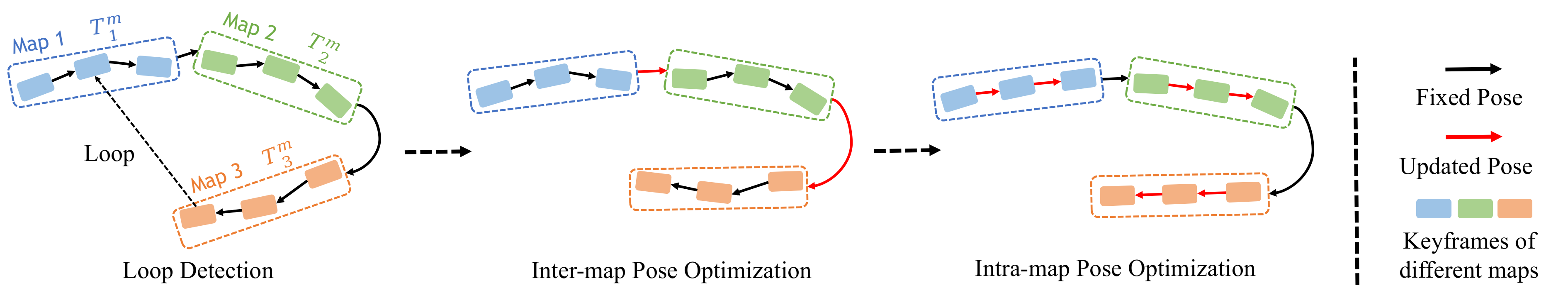}    
\caption{Illustration of hierarchical pose optimization. The optimization process consists of two steps: Inter-map pose optimization and intra-map pose optimization. When loop closure happens, we first perform inter-map optimization to update the pose of each map, $\{T_{i}^m\}$. Then, perform global bundle adjustment of keyframes within a map.}
\vspace{-2mm}
\label{fig:hierar_pose_optim}
\end{figure*}

\subsection{Hierarchical Pose Optimization}
\label{sec:hierarchical_pose_optim} 
After a valid loop candidate is found, we perform a hierarchical pose optimization, which consists of two steps: We first optimize the relative pose between maps according to the loop closure or adjacent frames with visual overlap. We call it the inter-map optimization step. Then we perform global bundle adjustments for keyframes within a single map to optimize the poses of keyframes, which is the intra-map optimization step. The whole process is shown in ~\autoref{fig:hierar_pose_optim}. 

\vspace{1mm}
\noindent\textbf{Inter-map Optimization:} When finding loop closure, we first optimize the relative pose between maps.
During the reconstruction process, given the current frame $I_c$ in the source map, $M_{src}$, which is a loop closure in the target map, $M_{tar}$, we can perform inter-map pose optimization to update the pose of $M_{src}$ and $I_c$.
We first transform $I_c$ to the coordinate system in the target map with the following equation:
\begin{equation}
    T_{tar}^{c} = (T_{tar}^{m})^{-1} \cdot T^{m}_{src} \cdot T^{c}_{src}
\end{equation}
where $T_{tar}^{c}$/$T_{src}^{c}$ are the pose of frame $I_c$ in the target/source map coordinate system, respectively. And $T_i^m$ represents the pose of the map $M_i$ in the world system. 

Then, we sample rays from pose $T_{tar}^{c}$ and compute loss between rendered values and ground truth observation with the following equation:
\begin{equation}
    \min_{T_{src}^m,T_{src}^c} \mathcal{L}(\sum_{i}F_{\theta}(\text{TriLerp}(T_{tar}^{c}\mathbf{p}_i, \mathbf{e})), C^{gt}, D^{gt})
\end{equation}
where $\mathcal{L}(\cdot)$ is the optimization objectives in~\autoref{sec:render}.

\vspace{1mm}
\noindent\textbf{Intra-map Optimization:} Once we have updated the relative poses between different maps, we start the global bundle adjustment within each of the two matched maps. Intra-map optimization is similar to the mapping process we explained earlier in~\autoref{sec:mapping}. The difference is that we update the poses of all keyframes for loop optimization instead of updating keyframes of a fixed local window size during the mapping process. Since the pose drifts within a single map have already reached a relatively good local minimum during the single-map mapping process. So, we just need to fine-tune the pose of keyframes for the consistency of each sub-map. As described above, we decompose the pose optimization of large scenes into different segments of pose optimization and use a hierarchical approach to perform bundle adjustment to reduce the pose drift.
\section{Experiments}
\label{sec:exp}

\subsection{Experimental Setup}
\label{subsec:setup}

\noindent\textbf{Datasets}: 
In the experiments, we evaluate our proposed Vox-Fusion++ system using two publicly available datasets and two self-captured scenes: 
(1) Replica dataset~\cite{julian:2019:replica}: This dataset contains $18$ different sequences captured by a camera rig. 
Following the setting in~\cite{sucar:2021:imap}, we use $8$ rendered sequences in our experiments.
(2) ScanNet dataset~\cite{dai:2017:scannet}: This dataset contains over $1,000$ captured RGB-D sequences along with ground truth poses estimated from a SLAM system~\cite{dai:2017:bundlefusion}. Following~\cite{zhu:2021:niceslam,sucar:2021:imap}, we leverage this dataset to assess the performance of our system in various scenarios.
(3) Two large-scale indoor scenes (\textit{scene 01} and \textit{scene 02}) that we captured in weakly textured and geometrically structured regions. 
Those scenes are captured using a handheld Azure Kinect camera, which comprises high-quality RGB and depth images, making them ideal for evaluating the capabilities in handling large-scale indoor environments.

\vspace{1mm}
\noindent\textbf{Evaluation Metrics}: 
In our evaluation, we employ various metrics to assess the performance of our Vox-Fusion++ system and compare it with other competing methods. 
Similar to previous works~\cite{sucar:2021:imap, zhu:2021:niceslam}, we focus on reconstruction quality and measure accuracy and completion.
The mesh accuracy (Acc.) is quantified using the unidirectional Chamfer distance between the reconstructed mesh and the ground truth. 
The completion (Comp.) metric, on the other hand, measures the distance the other way around. 
Additionally, we compute the completion ratio (Comp. Ratio), which represents the percentage of reconstructed points whose distance to the ground truth mesh is smaller than $5cm$.
The Chamfer distance between two mesh is formulated as follows:
\begin{equation}
\label{eqn:chamfer}
\begin{split}
    D_{Chamfer} = |P|^{-1} \sum_{(p,q) \in \Lambda_{P,Q}} \Vert p-q \Vert^2, \\
    \Lambda_{Q,P}^* = \{(p, argmin_{q} \Vert p - q \Vert)\}.
\end{split}
\end{equation}
Here, $P$ and $Q$ are two point sets sampled from the reconstructed and ground truth meshes, respectively.

For the evaluation of pose estimation, we use the widely-used absolute trajectory error (ATE) metric, calculated with the scripts provided by~\cite{sturm:2012:tumrgbd}.
The ATE measures the absolute translational difference between the estimated camera trajectory and the ground truth trajectory.

\vspace{1mm}
\noindent\textbf{Implementation Details}: 
Our encoder network is implemented as an MLP comprising several fully connected layers (FC). 
The input to the network is a $16$-$D$ feature vector interpolated from voxel embedding. 
The features are generally processed by $2$-$4$ FC layers that each have $128$ hidden units. 
The SDF decoder head outputs a scalar SDF value $s$ and a $128$-$D$ feature vector. 
The color decoder head has two FC layers with $256$ hidden units each and finally outputs $3$ dimension radiance value.
We apply the sigmoid function on the radiance value to generate RGB color in the range $[0, 1]$. 
The step size ratio for sampling points inside a voxel is generally set to $0.05$-$0.1$. 
For all scenes, we utilize a fixed voxel size of $0.2$ meters.

\begin{figure*}[tbp]
  \centering
  \scriptsize
  \setlength{\tabcolsep}{0.5pt}
  \newcommand{\sz}{0.28}
  \begin{tabular}{c@{\hspace{3pt}}c@{\hspace{3pt}}c@{\hspace{3pt}}c}
  
  \makecell{\includegraphics[height=\sz\columnwidth]{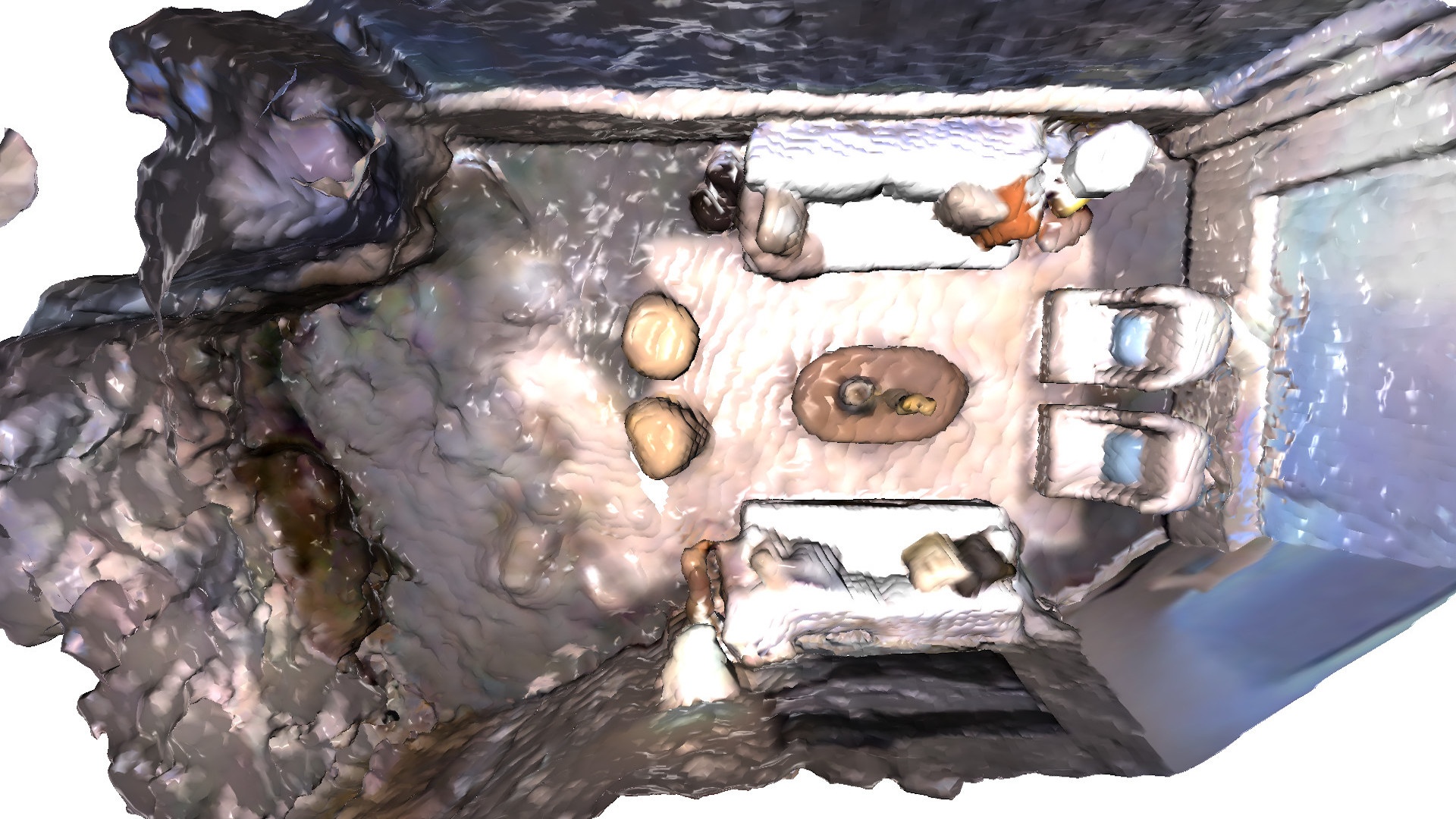}} & 
  \makecell{\includegraphics[height=\sz\columnwidth]{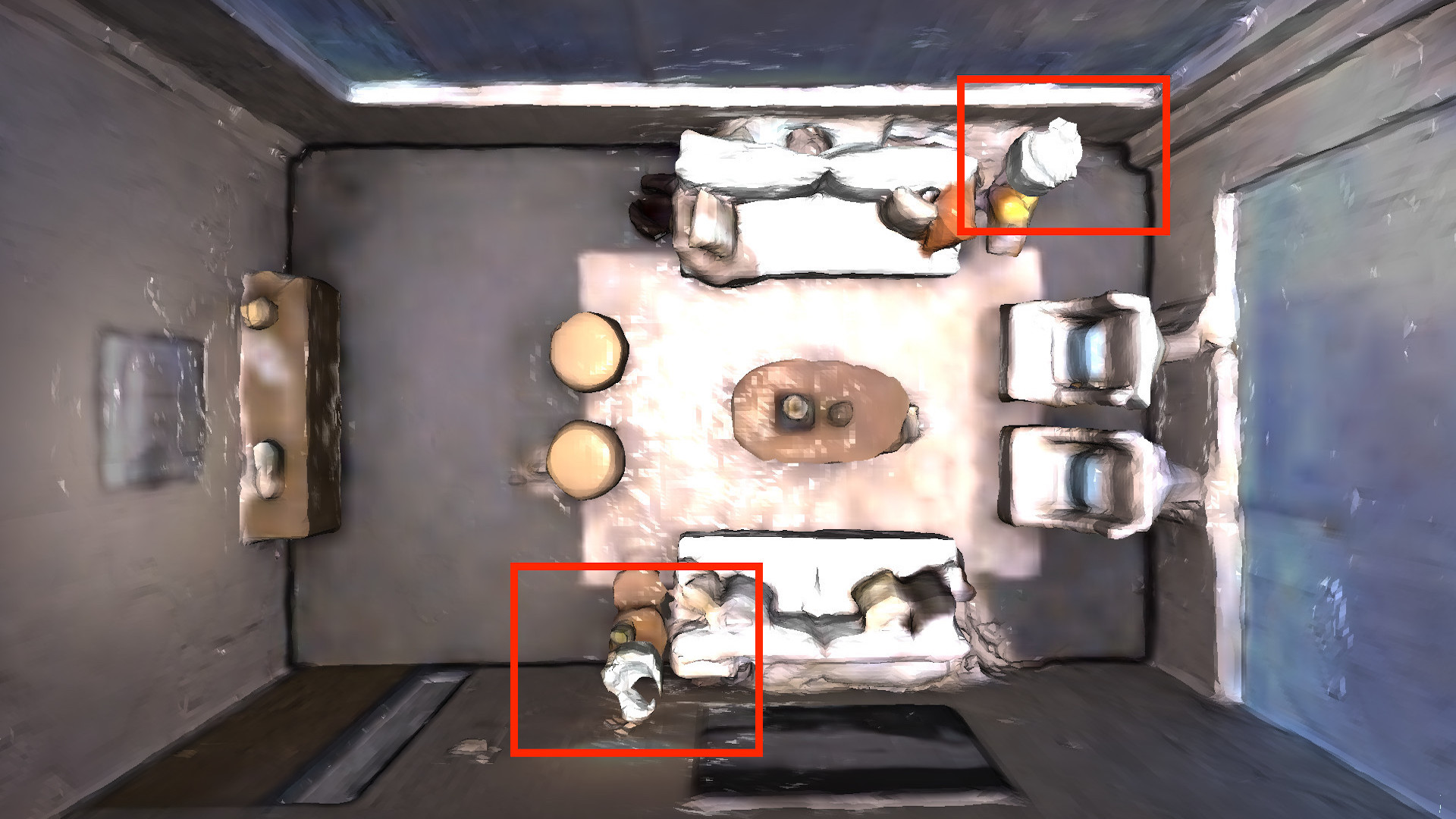}} & 
  \makecell{\includegraphics[height=\sz\columnwidth]{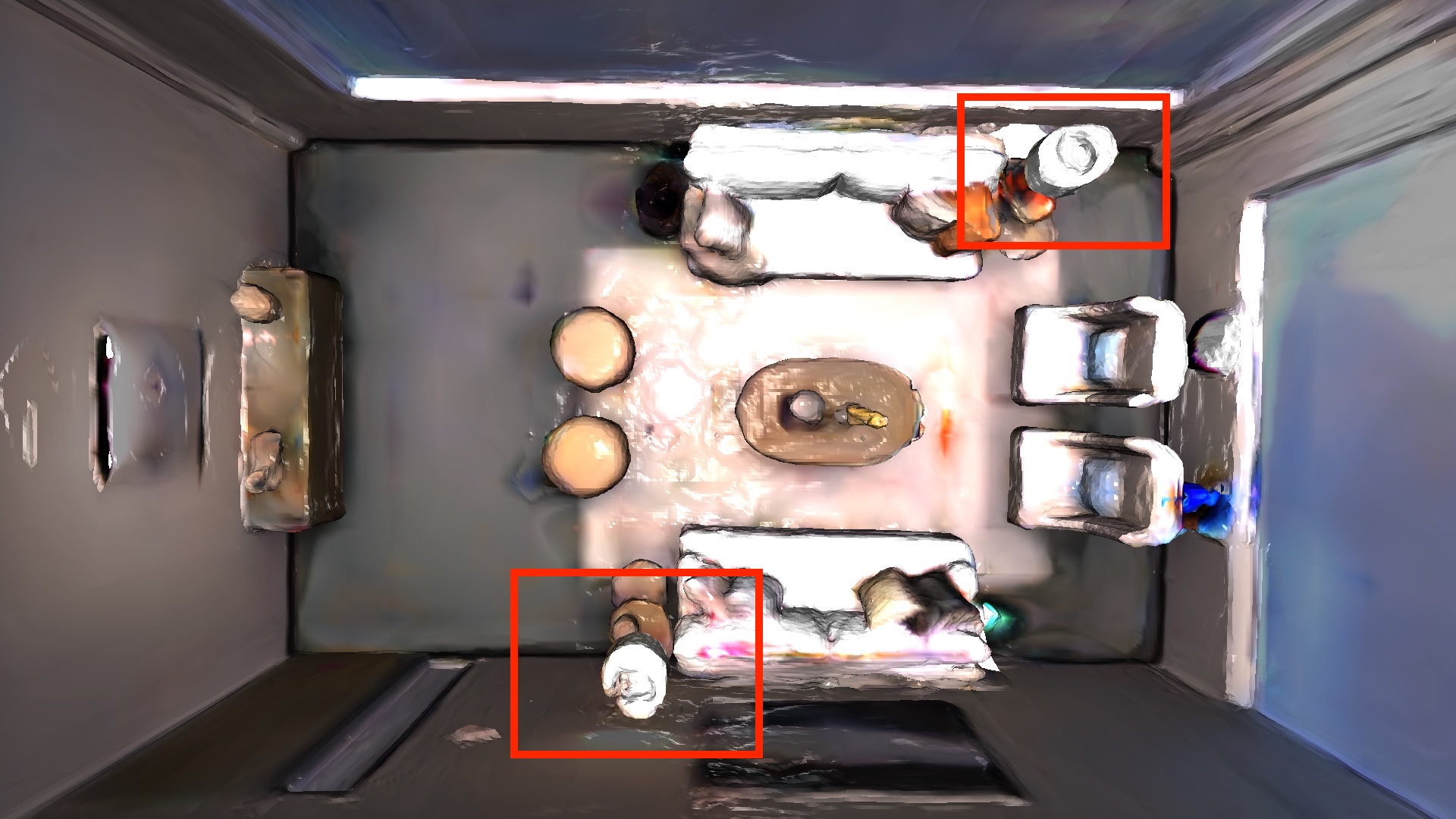}} & 
  \makecell{\includegraphics[height=\sz\columnwidth]{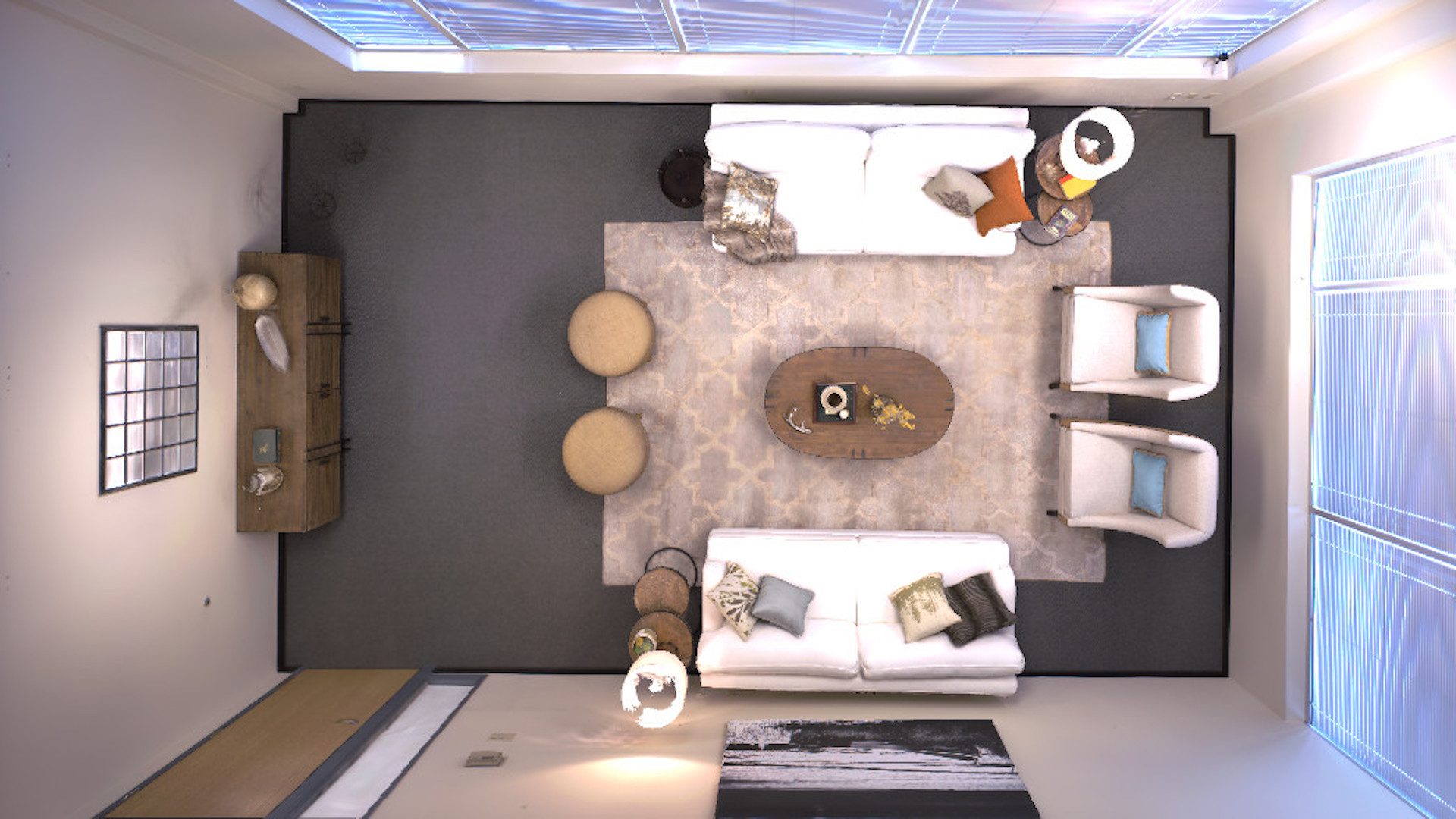}} \\

  \makecell{\includegraphics[height=\sz\columnwidth]{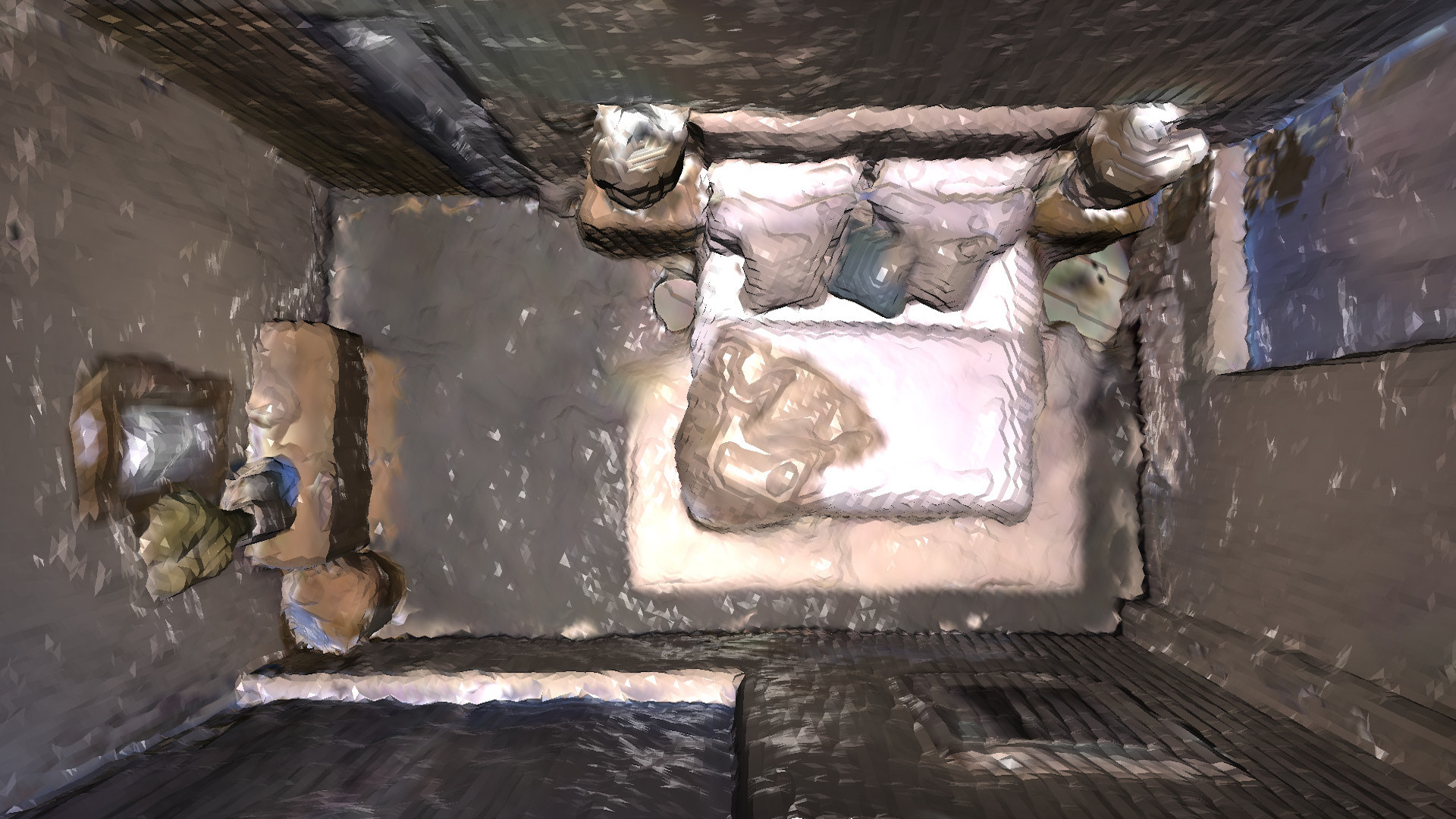}} & 
  \makecell{\includegraphics[height=\sz\columnwidth]{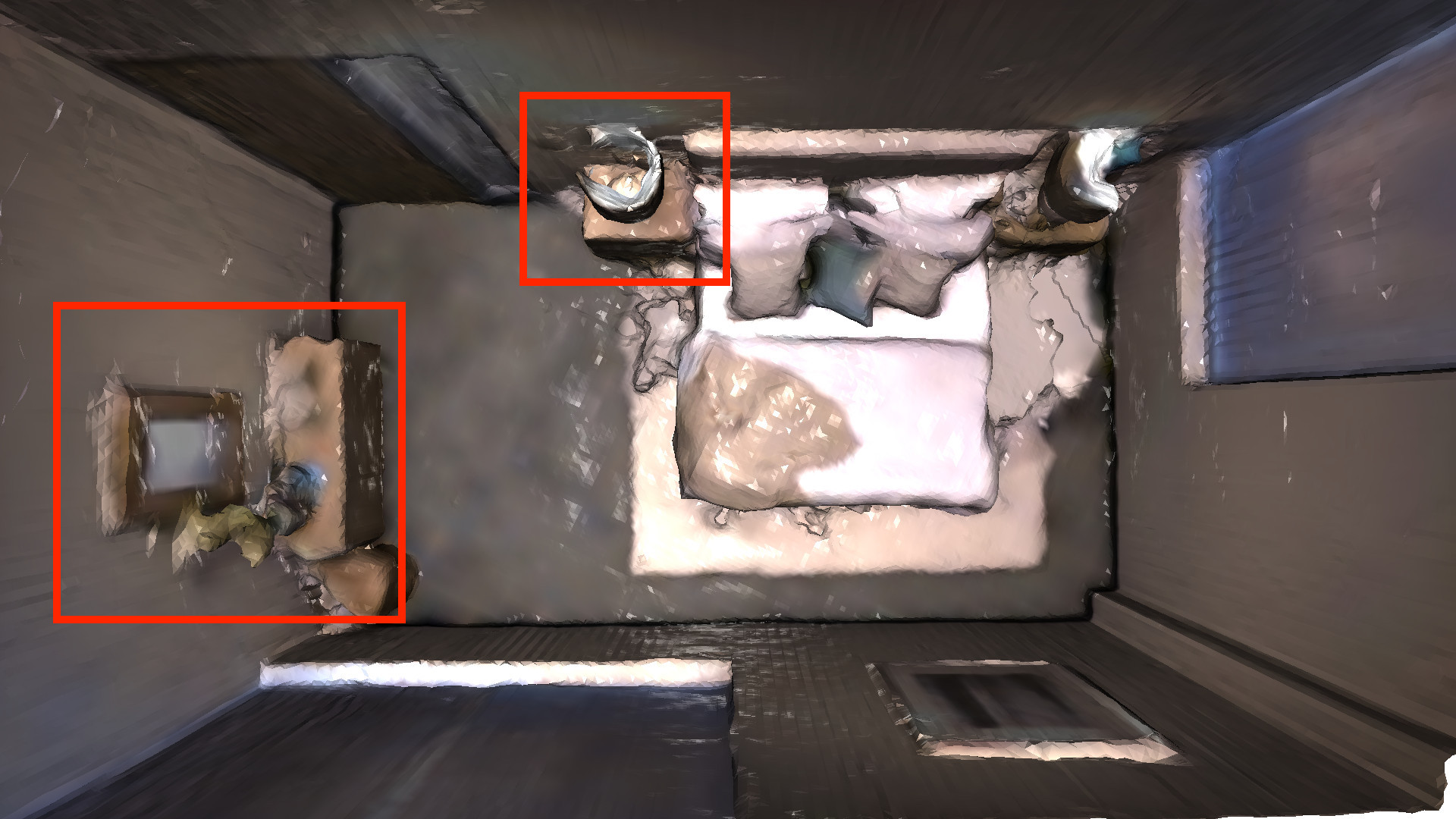}} & 
  \makecell{\includegraphics[height=\sz\columnwidth]{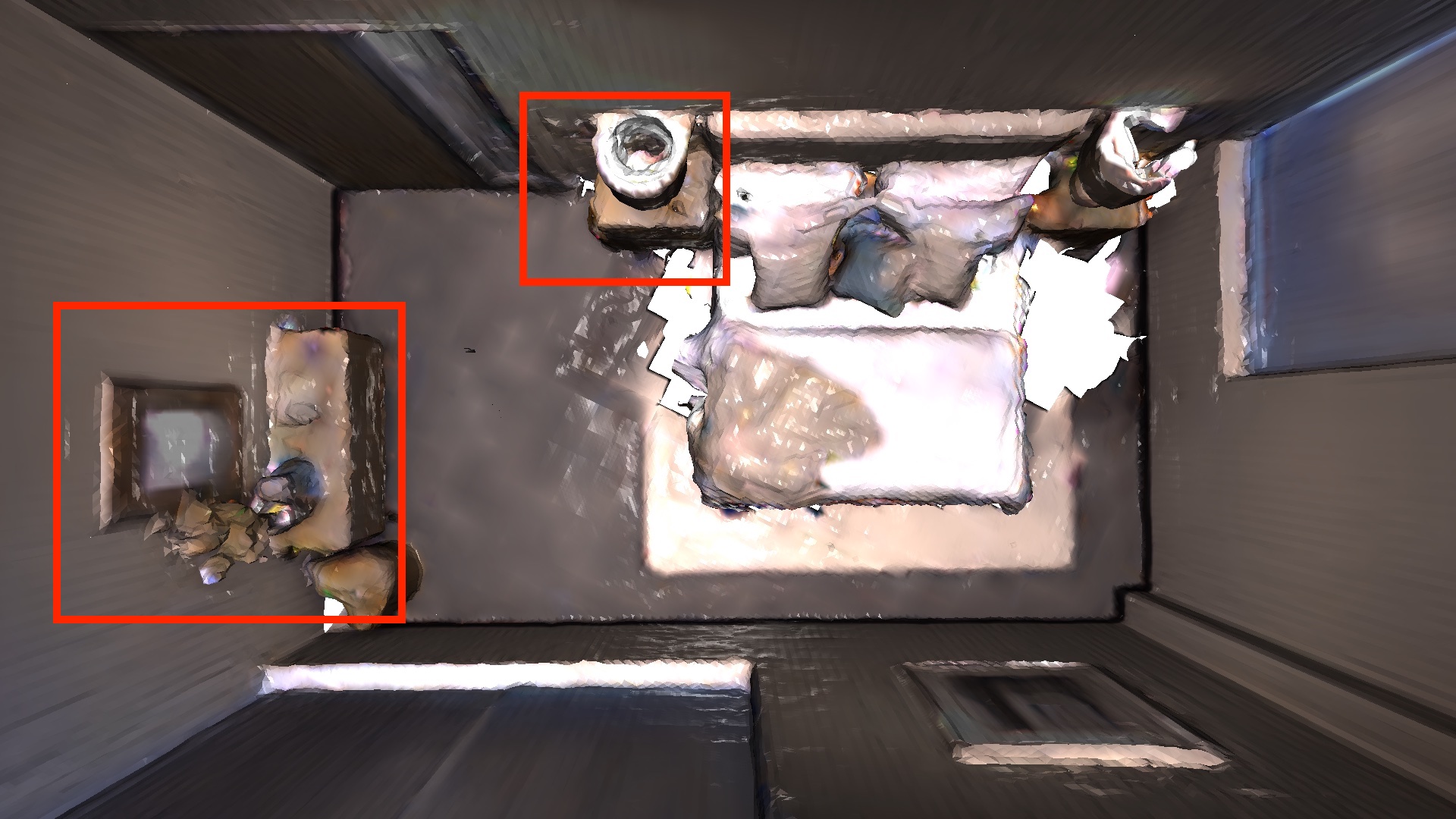}} & 
  \makecell{\includegraphics[height=\sz\columnwidth]{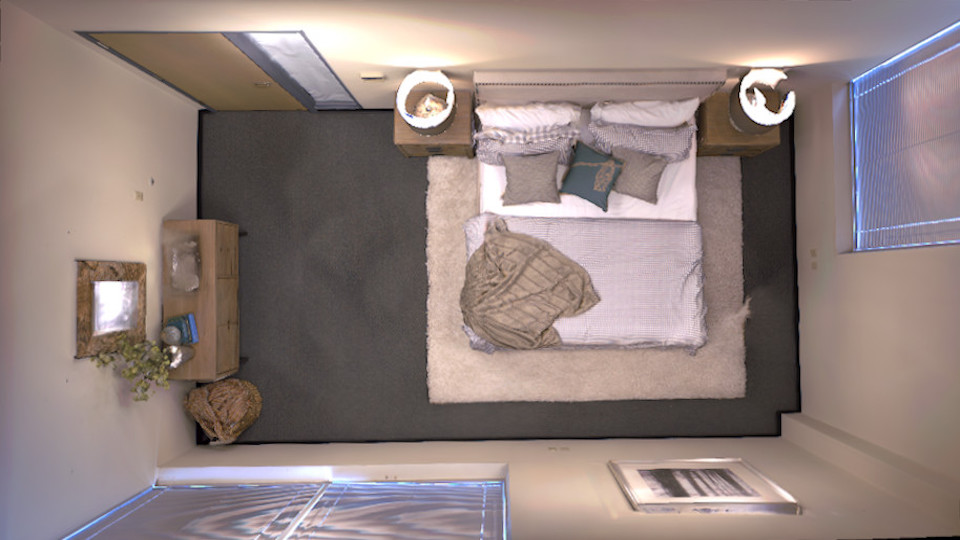}} \\

  \makecell{\includegraphics[height=\sz\columnwidth]{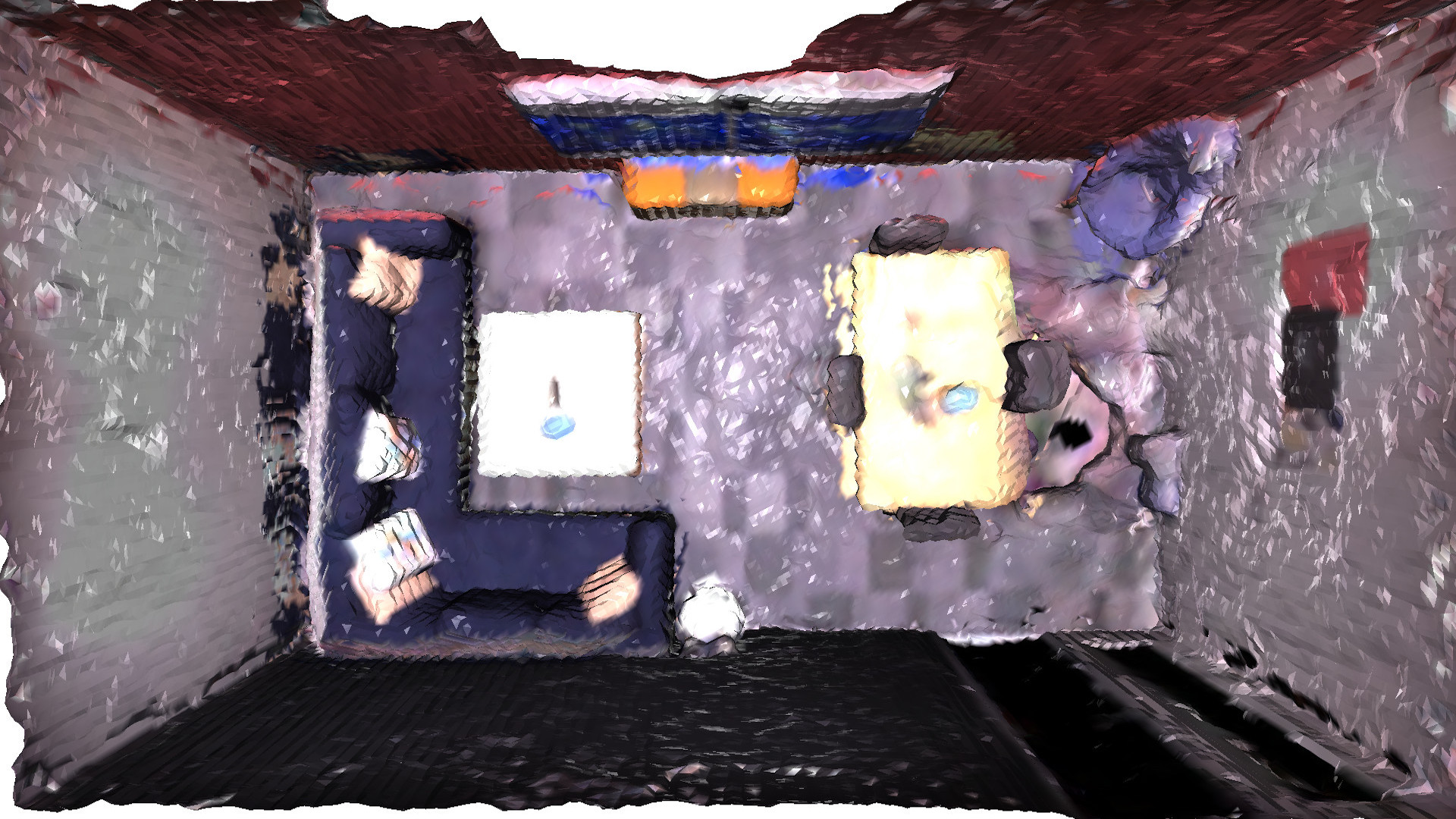}} & 
  \makecell{\includegraphics[height=\sz\columnwidth]{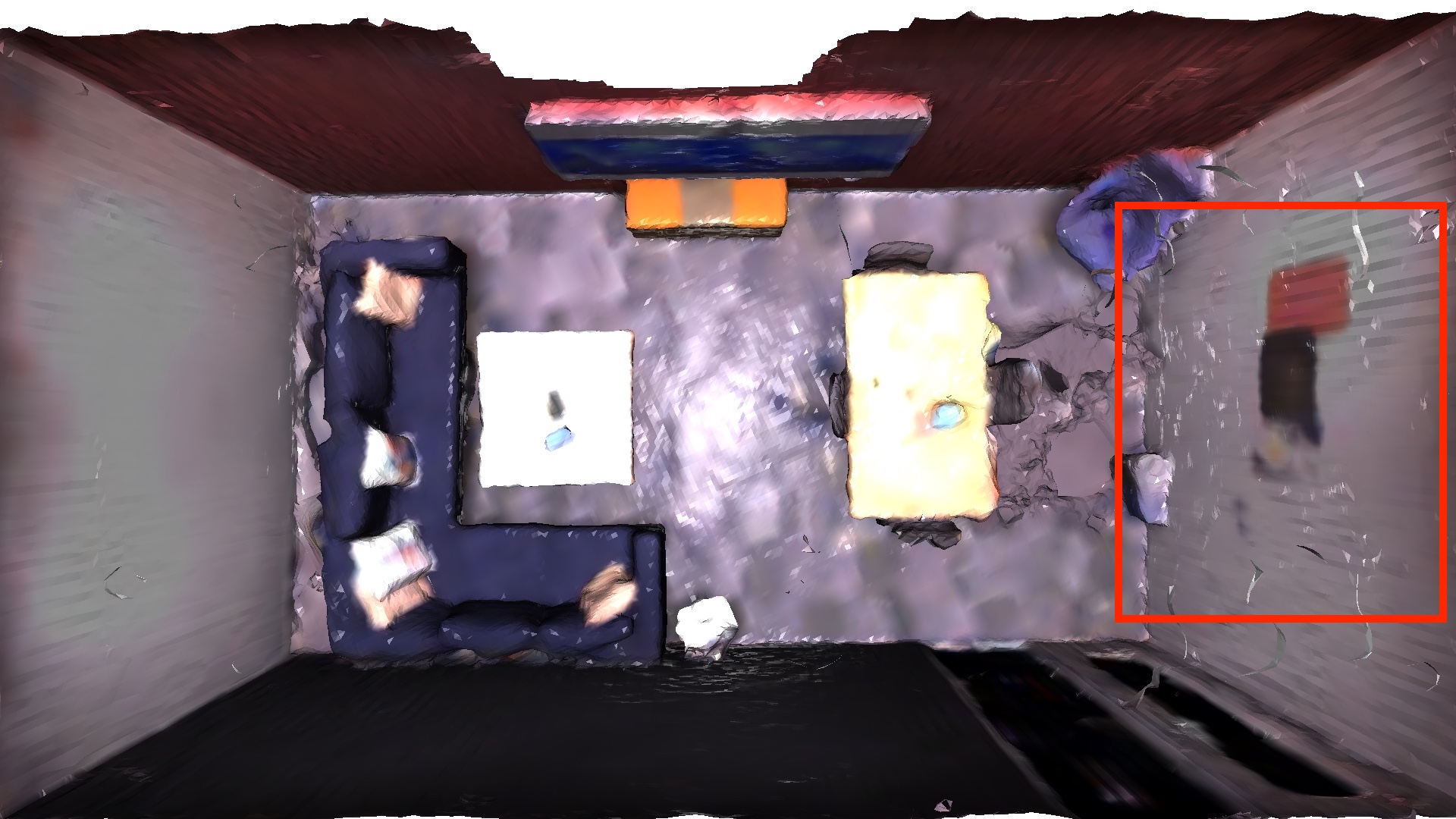}} & 
  \makecell{\includegraphics[height=\sz\columnwidth]{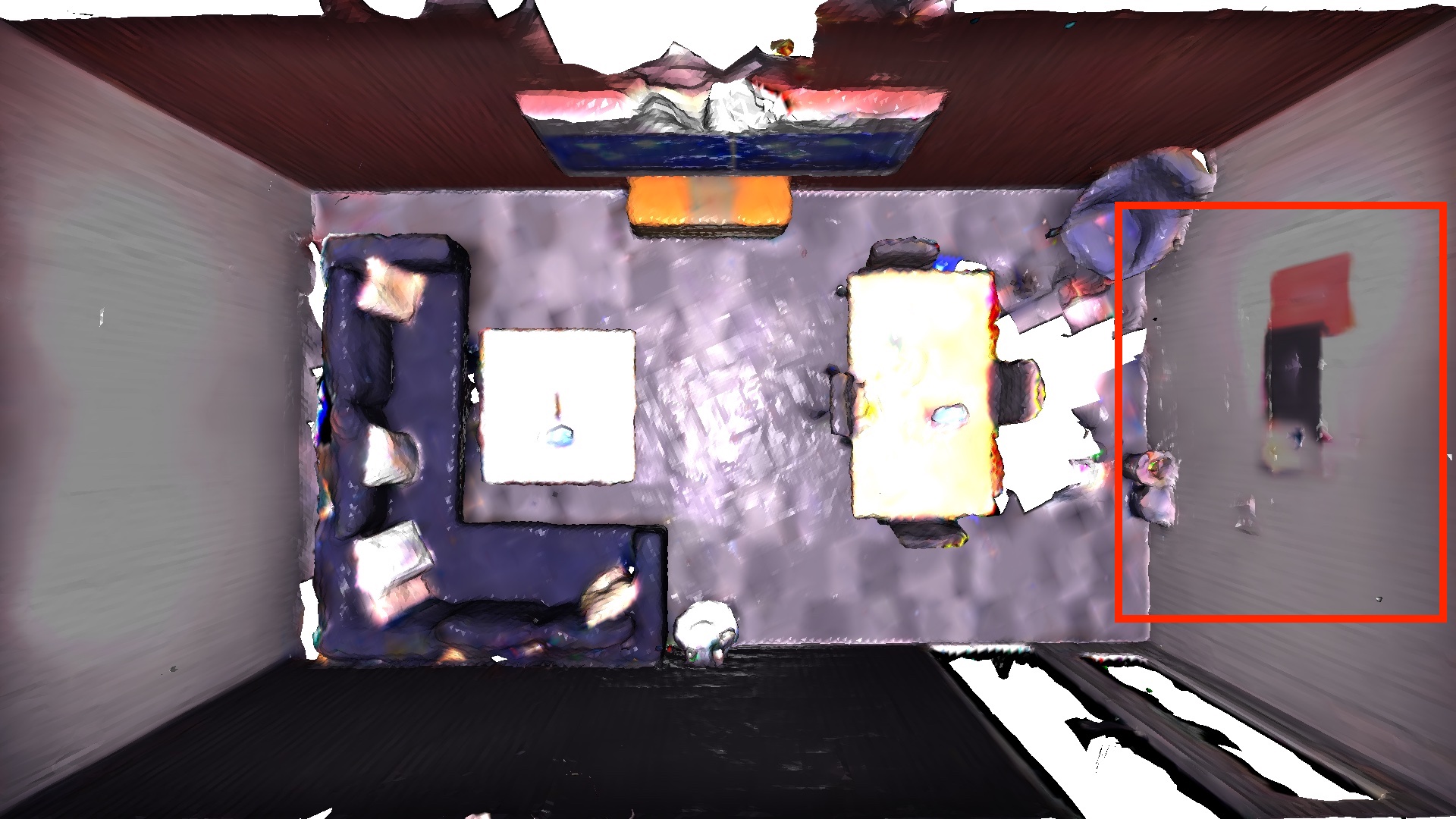}} & 
  \makecell{\includegraphics[height=\sz\columnwidth]{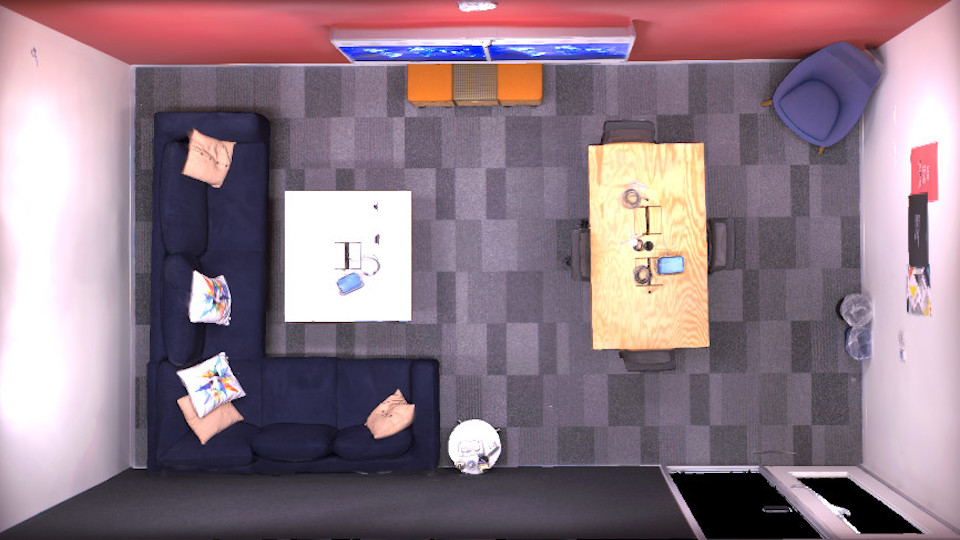}} \\

  \makecell{\includegraphics[height=\sz\columnwidth]{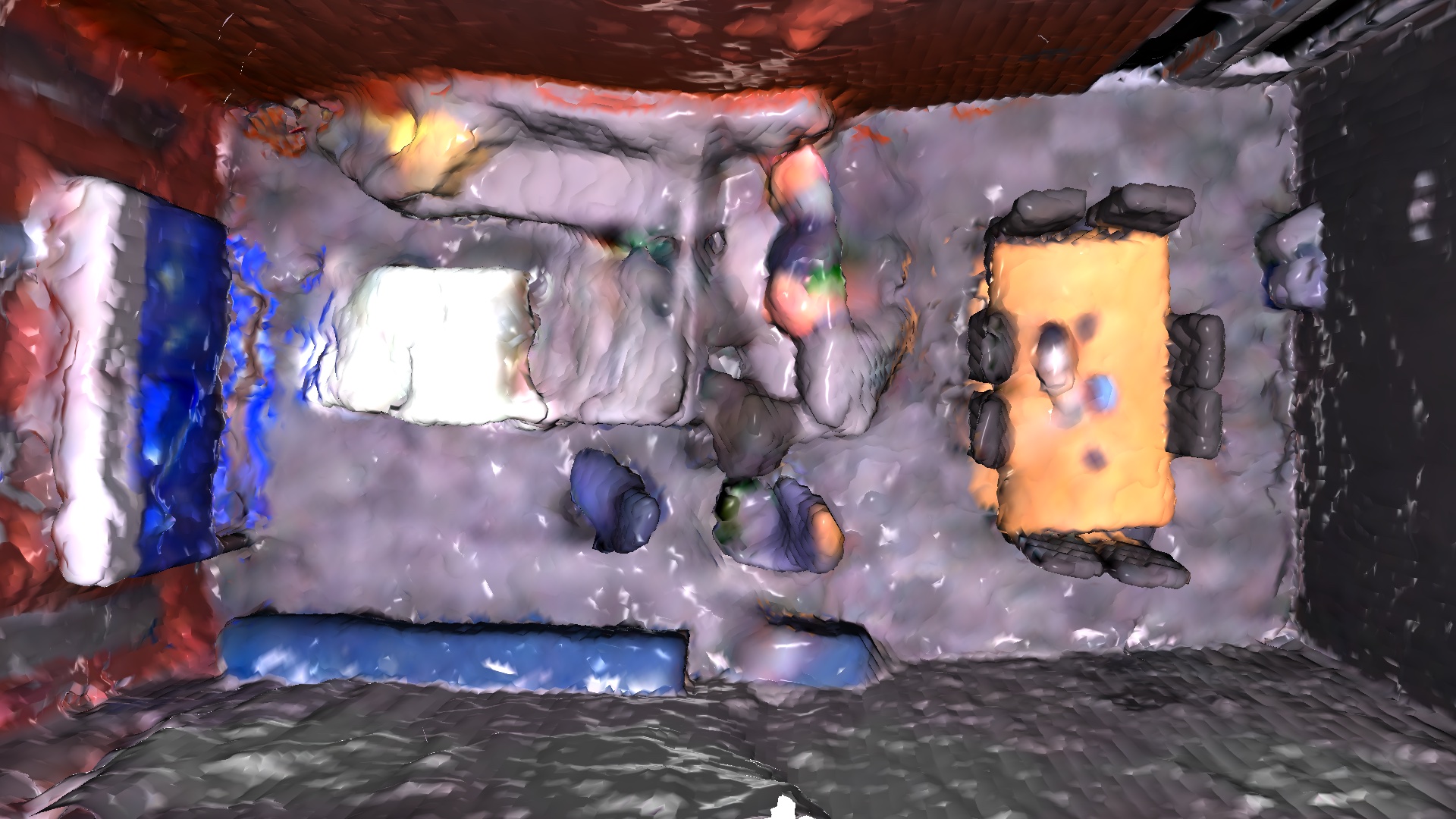}} & 
  \makecell{\includegraphics[height=\sz\columnwidth]{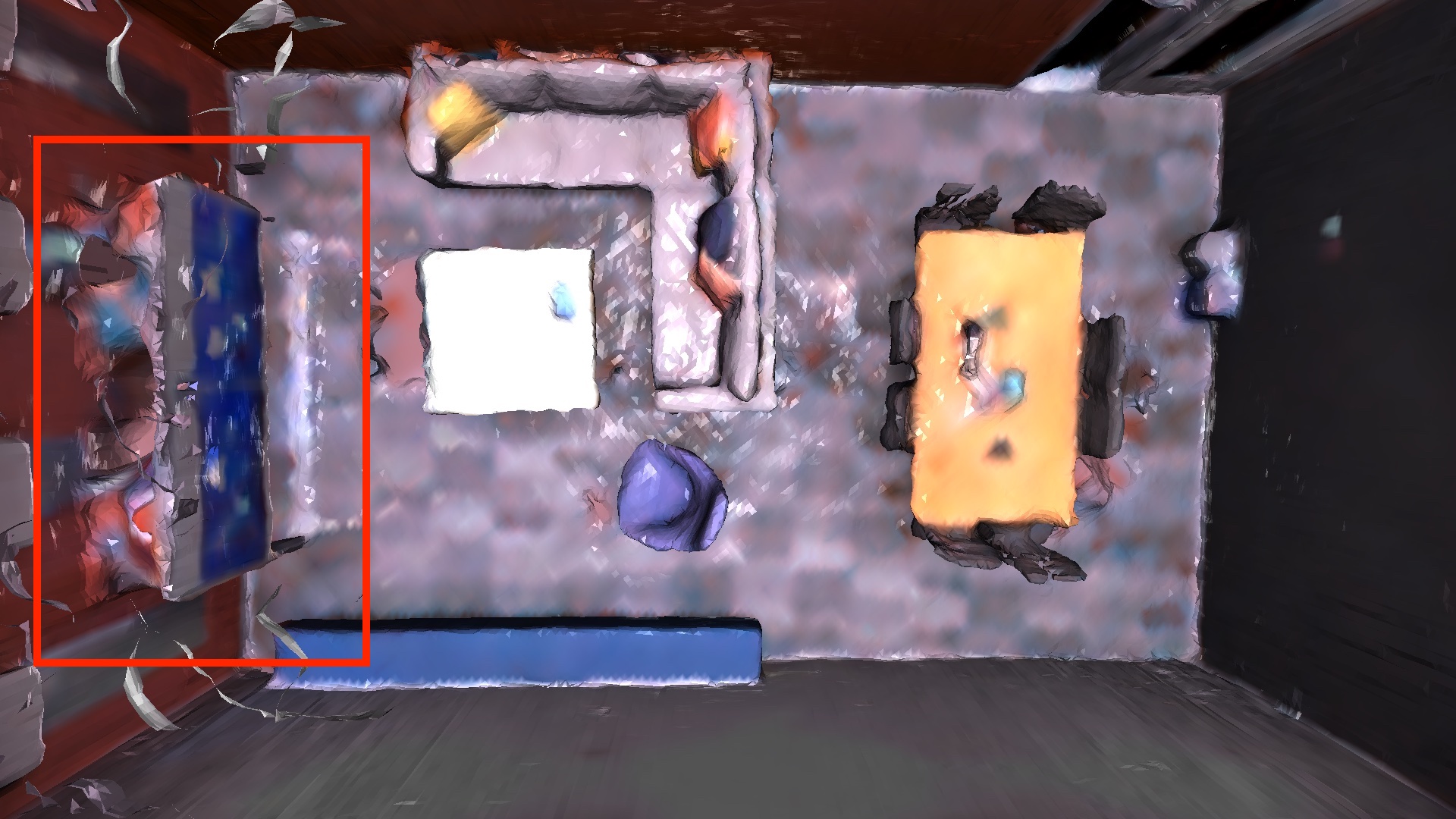}} & 
  \makecell{\includegraphics[height=\sz\columnwidth]{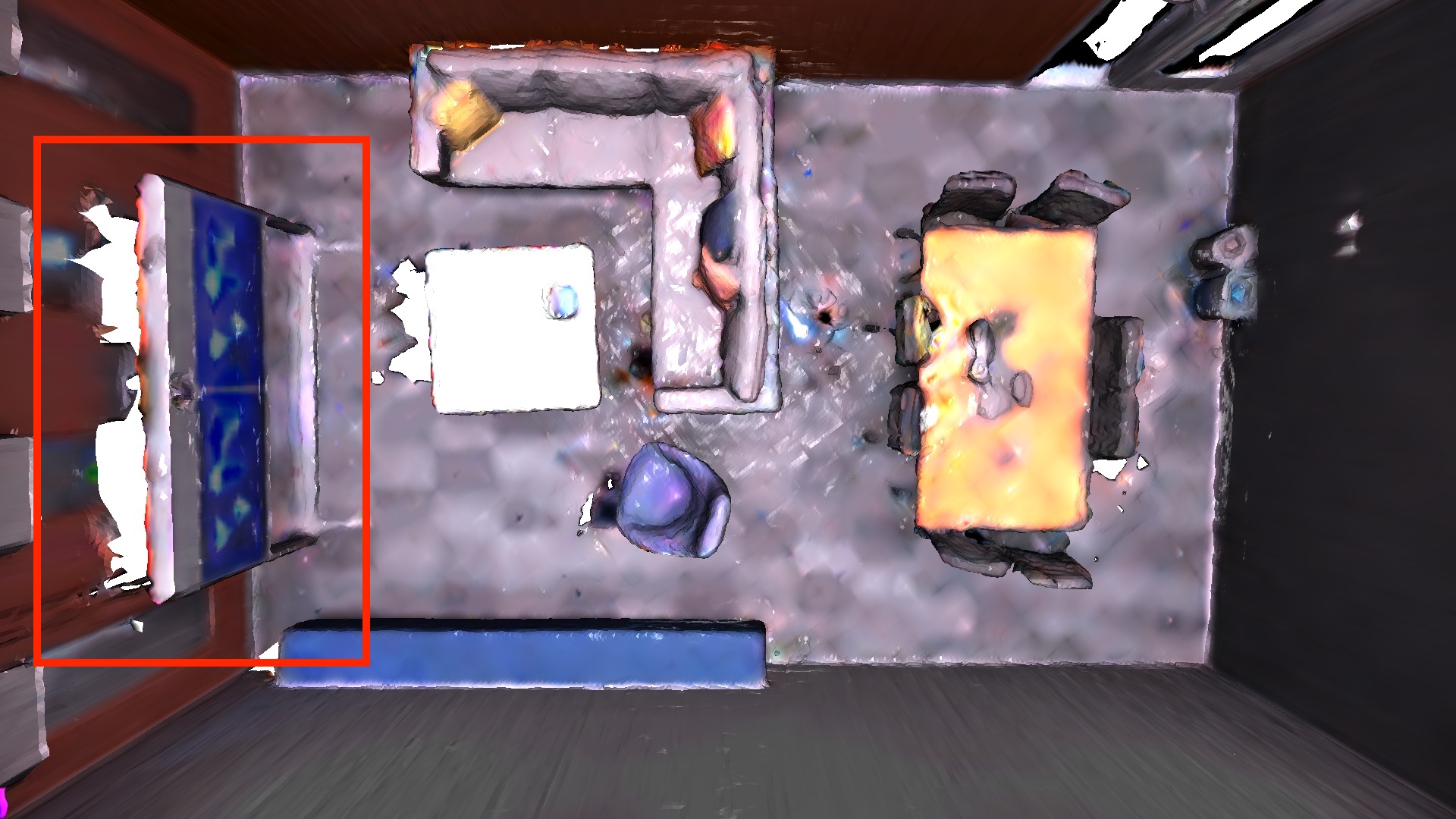}} & 
  \makecell{\includegraphics[height=\sz\columnwidth]{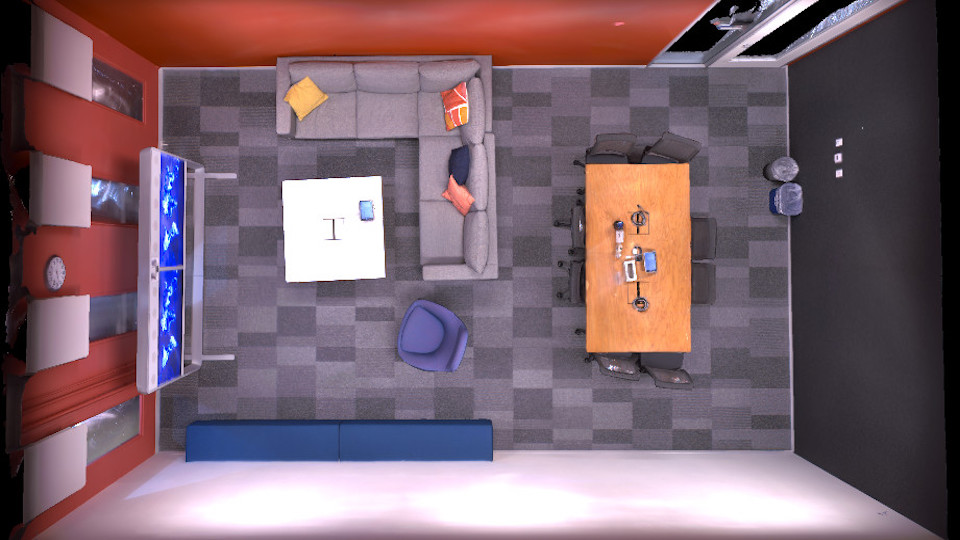}} \\

  \makecell{\tt \footnotesize{\text{iMap$^*$~\cite{sucar:2021:imap}}}} & \makecell{\tt \footnotesize{NICE-SLAM~\cite{zhu:2021:niceslam}}} & \makecell{\tt \footnotesize{Ours}} & \makecell{\tt \footnotesize{GT}}\\
  \end{tabular}
  \caption{Qualitative reconstruction results on the Replica dataset. From left to right, we show the results of scene reconstruction of different methods (iMAP$^{*}$, NICE-SLAM, our method, and ground truth). It can be clearly seen that our reconstruction results are much better than iMAP$^{*}$. To better show the difference in reconstruction between NICE-SLAM and our method, we use red boxes in the figures to indicate the improvements over NICE-SLAM.}
    \label{fig:replica}
  \vspace{-5pt}
\end{figure*}
\begin{table*}
  \caption{Trajectory estimation results of 8 scenes in the Replica dataset~\cite{julian:2019:replica}. Compared with iMap~\cite{sucar:2021:imap}, NICE-SLAM~\cite{zhu:2021:niceslam} and Co-SLAM~\cite{co-slam}, our method obtained better results on all sequences.}
  \label{tab:replica_ate}
  \scriptsize%
	\centering%
  \begin{tabular}{%
	l%
	l%
	*{2}{c}%
	*{2}{c}%
	*{2}{c}%
	*{2}{c}%
	*{2}{c}%
	*{2}{c}%
	*{2}{c}%
	*{2}{c}%
	*{2}{c}%
	}
  \toprule
  Methods & Metric & Room-0 & Room-1 & Room-2 & Office-0 & Office-1 & Office-2 & Office-3 & Office-4 & Avg. \\
  \midrule 
  \multirow{3}{*}{iMap*~\cite{sucar:2021:imap}} & \textbf{RMSE}[m]$\downarrow$ & 0.7005 & 0.0453 &0.0220 & 0.0232 & 0.0174 & 0.0487 & 0.5840 & 0.0262 & 0.1834 \\
  & \textbf{mean}[m]$\downarrow$ & 0.5891 & 0.0395 & 0.0195 & 0.0165 & 0.0155 & 0.0319  & 0.5488  & 0.0215  & 0.1603 \\
  & \textbf{median}[m]$\downarrow$ & 0.4478 & 0.0335 & 0.0173 & 0.0135 & 0.0137 & 0.0235  & 0.4756  & 0.0186  & 0.1304 \\
  \midrule
  \multirow{3}{*}{NICE-SLAM~\cite{zhu:2021:niceslam}} & \textbf{RMSE}[m]$\downarrow$ & 0.0169 & 0.0204 & 0.0155 & 0.0099 & 0.0090 & 0.0139 & 0.0397 & 0.0308 & 0.0195 \\
  & \textbf{mean}[m]$\downarrow$ & 0.0150 & 0.0180  & 0.0118  & 0.0086 & 0.0081 & 0.0120  & 0.0205 & 0.0209 & 0.0144 \\
  & \textbf{median}[m]$\downarrow$ & 0.0138 & 0.0167 & 0.0098  & 0.0076  & 0.0074  & 0.0109  & 0.0128  & 0.0153  & 0.0118 \\
  \midrule
  \multirow{3}{*}{Co-SLAM~\cite{co-slam}} & \textbf{RMSE}[m]$\downarrow$ & 0.0077 & 0.0104 & 0.0109 & 0.0058 & 0.0053 & 0.0205 & 0.0149 & 0.0084 & 0.0099 \\
  & \textbf{mean}[m]$\downarrow$ & 0.0066 & 0.0074  & 0.0092  & 0.0048 & 0.0046 & 0.0186  & 0.0140 & 0.0073 & 0.0091 \\
  & \textbf{median}[m]$\downarrow$ & 0.0057 & 0.0057 & 0.0084  & 0.0041  & 0.0041  & 0.0158  & 0.0136  & 0.0064  & 0.0080 \\
  \midrule
  \multirow{3}{*}{Ours} & \textbf{RMSE}[m]$\downarrow$ & \textbf{0.0038} & \textbf{0.0047} & \textbf{0.0049} & \textbf{0.0044} & \textbf{0.0042} & \textbf{0.0062} & \textbf{0.0041} & \textbf{0.0059} & \textbf{0.0048} \\
  & \textbf{mean}[m]$\downarrow$ & \textbf{0.0034} & \textbf{0.0040} & \textbf{0.0041} & \textbf{0.0036} & \textbf{0.0038} & \textbf{0.0055} & \textbf{0.0037} & \textbf{0.0053} & \textbf{0.0042} \\
  & \textbf{median}[m]$\downarrow$ & \textbf{0.0031} & \textbf{0.0036} & \textbf{0.0038} & \textbf{0.0032} & \textbf{0.0036} & \textbf{0.0049} & \textbf{0.0035} & \textbf{0.0050} & \textbf{0.0038} \\
  \bottomrule
  \end{tabular}%
\end{table*}

\begin{table*}
  \caption{Reconstruction results of 8 scenes in the Replica dataset~\cite{julian:2019:replica}. Compared with iMap~\cite{sucar:2021:imap}, NICE-SLAM~\cite{zhu:2021:niceslam} and Co-SLAM~\cite{co-slam}, our approach consistently yields better results on all sequences.}
  \label{tab:recon_replica}
  \scriptsize%
	\centering%
  \begin{tabular}{%
	l%
	l%
	*{2}{c}%
	*{2}{c}%
	*{2}{c}%
	*{2}{c}%
	*{2}{c}%
	*{2}{c}%
	*{2}{c}%
	*{2}{c}%
	*{2}{c}%
	}
  \toprule
  Methods & Metric & Room-0 & Room-1 & Room-2 & Office-0 & Office-1 & Office-2 & Office-3 & Office-4 & Avg. \\
 \midrule 
  \multirow{3}{*}{iMap$^{*}$~\cite{sucar:2021:imap}} & \textbf{Acc.}[cm]$\downarrow$ & 3.58 & 3.69 & 4.68 & 5.87 & 3.71 & 4.81 & 4.27 & 4.83 & 4.43 \\
                          & \textbf{Comp.}[cm]$\downarrow$ & 5.06 & 4.87 & 5.51 & 6.11 & 5.26 & 5.65 & 5.45 & 6.59 & 5.56 \\
                          & \textbf{Comp. Ratio}[$<5$cm \%]$\uparrow$ & 83.91 & 83.45 & 75.53 & 77.71 & 79.64 & 77.22 & 77.34 & 77.63 & 79.06\\
  \midrule
  \multirow{3}{*}{NICE-SLAM~\cite{zhu:2021:niceslam}} & \textbf{Acc.}[cm]$\downarrow$ & 3.53 & 3.60 & 3.03 & 5.56 & 3.35 & 4.71 & 3.84 & 3.35 & 3.87 \\
                          & \textbf{Comp.}[cm]$\downarrow$ & 3.40 &  3.62 & 3.27 & 4.55 & 4.03 & 3.94 & 3.99 & 4.15 & 3.87 \\
                          & \textbf{Comp. Ratio}[$<5$cm \%]$\uparrow$ & 86.05 & 80.75 & 87.23 & 79.34 & 82.13 & 80.35 & 80.55 & 82.88 & 82.41 \\
  \midrule
  \multirow{3}{*}{Co-SLAM~\cite{co-slam}} & \textbf{Acc.}[cm]$\downarrow$ & 1.61 & 1.31 & 1.55 & 1.33 & 1.11 & 1.83 & 1.97 & 1.73 & 1.56 \\
                          & \textbf{Comp.}[cm]$\downarrow$ & 2.96 &  2.46 & 2.36 & 1.43 & 1.82 & 3.26 & 3.26 & 3.36 & 2.61 \\
                          & \textbf{Comp. Ratio}[$<5$cm \%]$\uparrow$ & 91.12 & 92.18 & 91.44 & 95.65 & 93.56 & 88.53 & 87.67 & 87.76 & 90.99 \\
  \midrule
  \multirow{3}{*}{Ours} & \textbf{Acc.}[cm]$\downarrow$ & \textbf{1.56} & \textbf{1.25} & \textbf{1.47} & \textbf{1.20} & \textbf{1.08} & \textbf{1.56} & \textbf{1.75} & \textbf{1.61} & \textbf{1.44} \\
                          & \textbf{Comp.}[cm]$\downarrow$ & \textbf{2.87} & \textbf{2.36} & \textbf{2.08} & \textbf{1.36} & \textbf{1.65} & \textbf{2.95} & \textbf{2.88} & \textbf{3.26} & \textbf{2.43} \\
                          & \textbf{Comp. Ratio}[$<5$cm \%]$\uparrow$ & \textbf{92.00} & \textbf{92.98} & \textbf{93.78} & \textbf{96.82} & \textbf{94.81} & \textbf{90.23} & \textbf{89.87} & \textbf{88.46} & \textbf{92.37} \\
  \bottomrule
  \end{tabular}%
\end{table*}
\subsection{Results on Replica Dataset}
We qualitatively compare our system with iMap~\cite{sucar:2021:imap}, NICE-SLAM~\cite{zhu:2021:niceslam}, and Co-SLAM~\cite{co-slam}. 
The trajectory estimation and reconstruction results are generated using their official codes. 
Subsequently, we present the trajectory estimation and surface reconstruction results.

\vspace{1mm}
\noindent\textbf{Trajectory Estimation: } 
We compare our trajectory estimation accuracy with three state-of-the-art neural SLAM systems~\cite{sucar:2021:imap, zhu:2021:niceslam,co-slam}. 
It is important to note that for the trajectory estimation experiment, we used the iMap implementation of~\cite{zhu:2021:niceslam} (denoted as iMap*).
The results are presented in~\autoref{tab:replica_ate}, and it shows that our system outperforms all baselines on three metrics. 
Moreover, we achieved significantly better results on camera pose estimation with a substantial margin. 
These results further validate that our system can produce state-of-the-art results on synthetic datasets.

\vspace{1mm}
\noindent\textbf{Surface Reconstruction:} 
In the surface reconstruction experiments, we show the qualitative results in~\autoref{fig:replica}. 
Comparing the reconstructed surfaces obtained by our Vox-Fusion++ system with iMap and NICE-SLAM, it is evident that our method produces superior meshes to iMap and performs comparably to NICE-SLAM.
One of the notable advantages of our approach lies in its handling of unobserved regions. 
While iMap and NICE-SLAM assume densely populated surfaces and create surfaces even in areas without observations, our explicit voxel map approach generates surfaces only within the visible sparse voxels. 
This enables us to achieve plausible hole fill-in effects while leaving large unobserved spaces empty, preventing discrepancies from occurring between the reconstruction and the ground truth.
By doing so, we effectively combine the best of both worlds. 
Although it might seem like a disadvantage at first, we argue that for real-world tasks it is often more important to know where has been observed and where has not.
In addition to qualitative results, we also quantitatively compared our system on reconstruction quality with iMap, NICE-SLAM, and Co-SLAM.
The results of reconstruction quality for iMap and NICE-SLAM are directly obtained from its paper~\cite{sucar:2021:imap,zhu:2021:niceslam}. The results for Co-SLAM are obtained with their released code.
To ensure a fair comparison, we used the same mesh culling approach~\cite{vox-fusion} for all methods. 
The results on reconstruction accuracy, as shown in~\autoref{tab:recon_replica}, demonstrate that our Vox-Fusion++ system outperforms three baselines across all metrics. 
These results further validate our observation that our system is capable of producing state-of-the-art results on synthetic datasets.

\begin{figure*}
  \centering
  \scriptsize
  \setlength{\tabcolsep}{0.5pt}
  \newcommand{\sz}{0.18}  %
  
  \begin{tabular}{c@{\hspace{3pt}}c@{\hspace{3pt}}c@{\hspace{3pt}}c@{\hspace{3pt}}c}
  \includegraphics[width=\sz\textwidth]{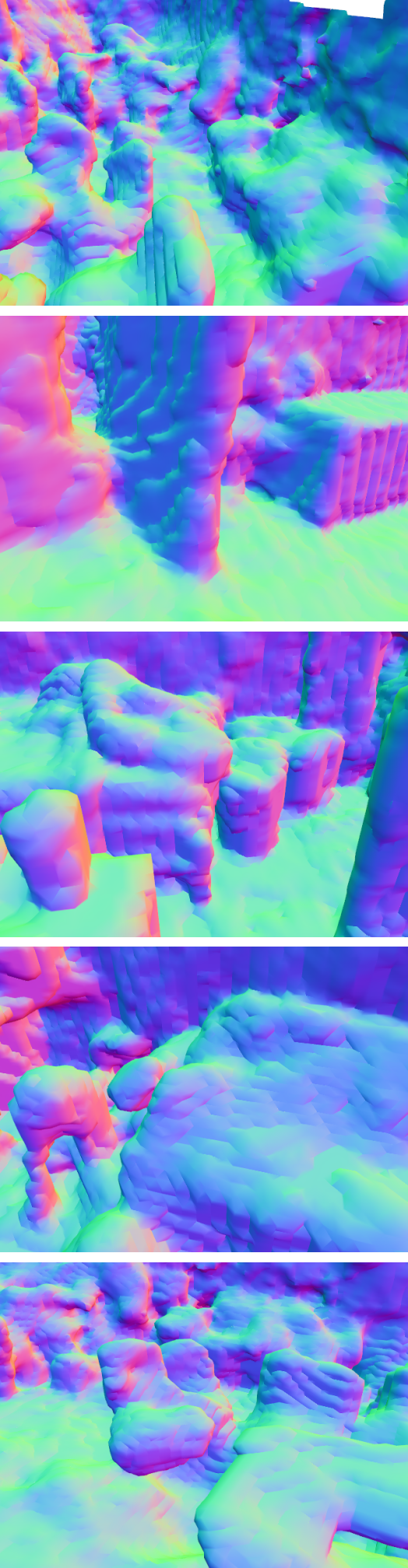} &
  \includegraphics[width=\sz\textwidth]{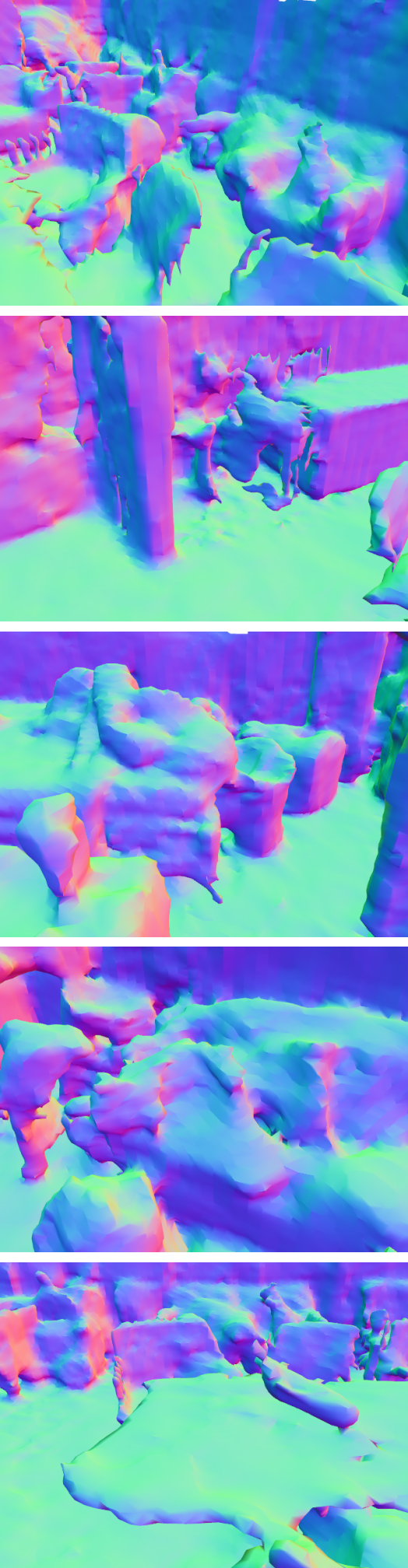} &
  \includegraphics[width=\sz\textwidth]{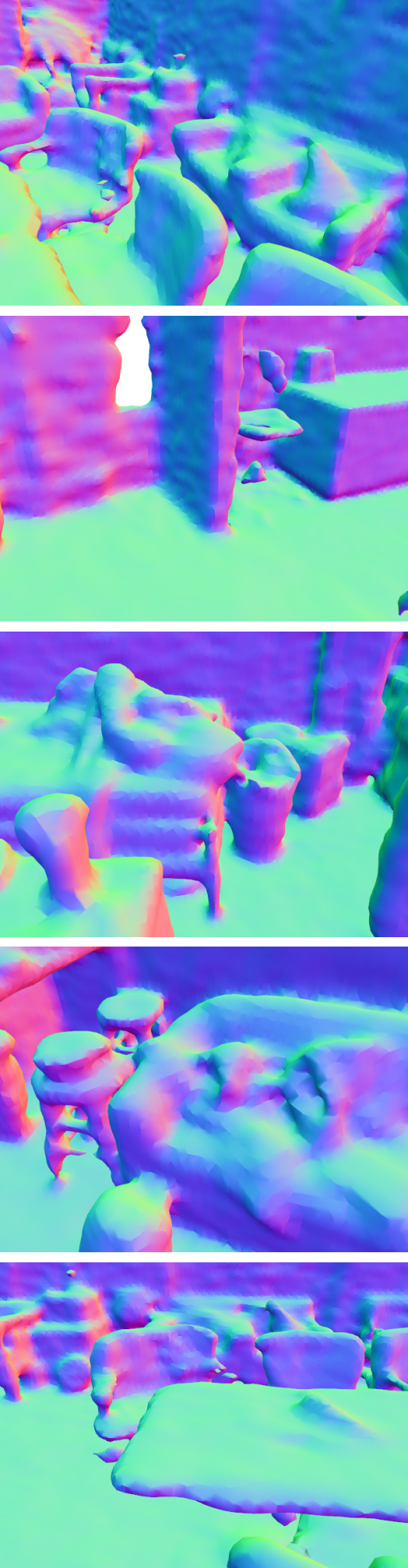} &
  \includegraphics[width=\sz\textwidth]{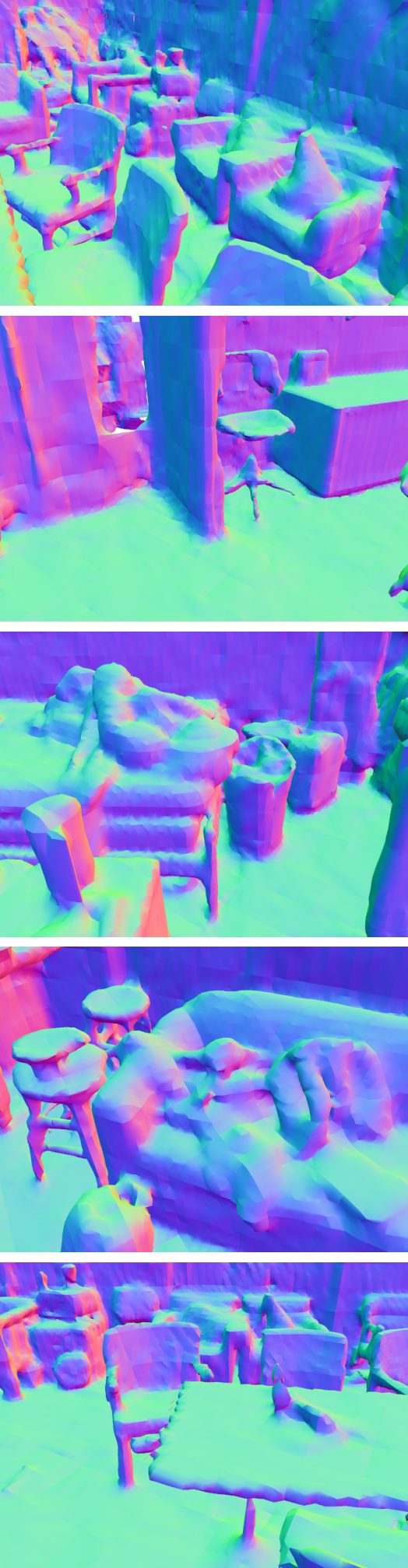} &
  \includegraphics[width=\sz\textwidth]{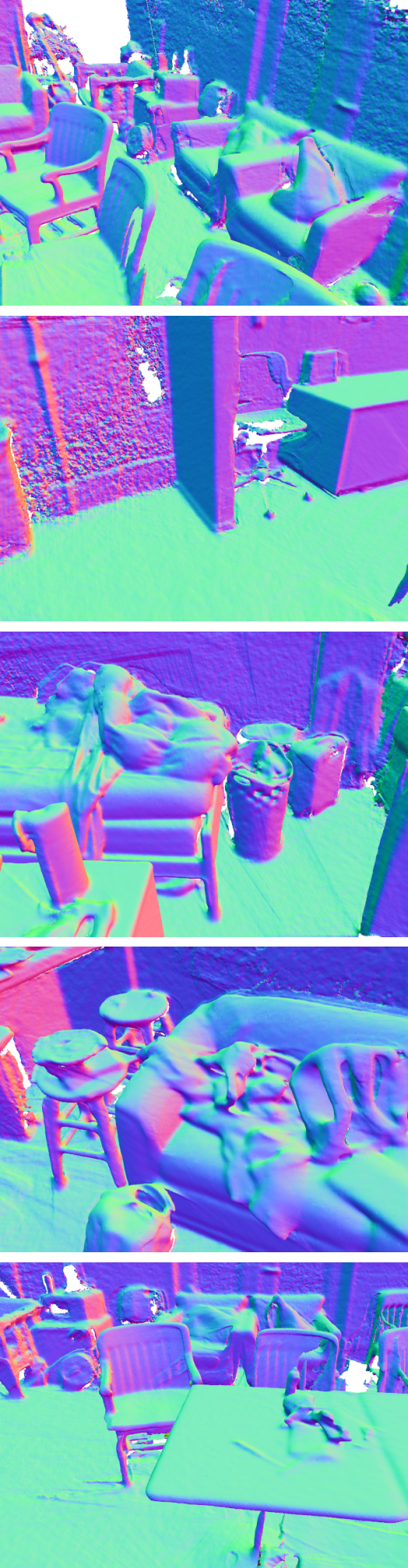} \\
  {\tt \footnotesize{iMap$^*$~\cite{sucar:2021:imap}}} & {\tt \footnotesize{NICE-SLAM~\cite{zhu:2021:niceslam}}} & {\tt \footnotesize{Co-SLAM~\cite{co-slam}}} & {\tt \footnotesize{Ours}} & {\tt \footnotesize{ScanNet Mesh}}\\
  \end{tabular}
\caption{Qualitative comparison on ScanNet dataset from different views. From left to right, we show the results of scene reconstruction of different methods (iMap$^*$, NICE-SLAM, Co-SLAM, ours, and ScanNet Mesh).}
    \label{fig:scannet-mesh-comp}
\end{figure*}

\subsection{Results on ScanNet Datasets}
Unlike synthetic datasets, real scans are noisier and contain erroneous measurements. 
Reconstructing real scans is considered a challenging task that has not yet been solved. 
We benchmarked our system on 6 selected sequences of ScanNet~\cite{dai:2017:scannet}. 
The selection of sequences is in line with~\cite{zhu:2021:niceslam}, and the results from DI-Fusion, iMap*, NICE-SLAM, and Co-SLAM are directly taken from~\cite{zhu:2021:niceslam,co-slam}. 
The surfaces of different views are obtained from reconstructed meshes generated by their officially released codes.

\begin{table}
  \scriptsize
  \centering%
  \newcommand{\sz}{0.5cm}
  \newcolumntype{P}[1]{>{\left}p{#1}}
  \caption{Trajectory estimation results (ATE RMSE (cm)$\downarrow$) on the ScanNet dataset of different methods.}
  \label{tab:scannet_ate}
  \begin{tabular}{p{1.9cm}p{\sz}p{\sz}p{\sz}p{\sz}p{\sz}p{\sz}p{\sz}}
  \toprule
  Scene ID & 0000 & 0059 & 0106 & 0169 & 0181 & 0207 & Avg.\\
  \midrule 
  DI-Fusion~\cite{huang:2021:difusion} & 62.99 & 128.00 & 18.50 & 75.80 & 87.88 & 100.19 & 78.89\\
  iMap*~\cite{sucar:2021:imap} & 55.95 & 32.06 & 17.50 & 70.51 & 32.10 & 11.91 & 36.67 \\
  NICE-SLAM~\cite{zhu:2021:niceslam} & 8.64 & 12.25 & 8.09 & 10.28 & 12.93 & 5.59 & 9.63 \\
  CO-SLAM~\cite{co-slam} & 7.13 & 11.14 & 9.36 & 5.90 & $\textbf{11.81}$ & 7.14 & 8.75 \\
  Ours & \textbf{6.38} & \textbf{7.28} & \textbf{6.75} & \textbf{5.86} & 13.68 & \textbf{4.73} & \textbf{7.44} \\
  \bottomrule
  \end{tabular}%
\end{table}

\begin{table}
  \scriptsize
  \centering%
  \newcommand{\sz}{0.6cm}
  \newcolumntype{P}[1]{>{\left}p{#1}}
  \caption{Trajectory estimation results (ATE RMSE (m)$\downarrow$) on our large indoor scenes of different methods.}
  \label{tab:loop_optimization}
  \begin{tabular}{p{1.8cm}p{\sz}p{\sz}p{\sz}p{\sz}p{\sz}p{\sz}}
  \toprule
  Scene ID & \multicolumn{3}{c}{\textit{Scene 01}} & \multicolumn{3}{c}{\textit{Scene 02}} \\
  Metric & RMSE & Mean & Median & RMSE & Mean & Median \\
  \midrule 
  W/O Loop ~\cite{vox-fusion} & 3.367 & 2.864 & 2.847 & 0.912 & 0.841 & 0.867 \\
  W Loop & \textbf{2.687} & \textbf{2.143} & \textbf{1.928} & \textbf{0.562} & \textbf{0.516} & \textbf{0.544} \\
  \bottomrule
  \end{tabular}%
\end{table}
\vspace{1mm}
\noindent\textbf{Trajectory Estimation:} The quantitative results are shown in~\autoref{tab:scannet_ate}.
It can be seen that despite the simplicity of our design, our method achieves better results than three baselines except for scene0181.
Please note that NICE-SLAM$/$Co-SLAM set the voxel size of the finest resolution to 16cm$/$4cm, but we use the voxel size of 20cm.
Although they use a finer voxel size to encode the scene, we still obtain better trajectory estimation results on most sequences.
Overall, our trajectory estimation results on the ScanNet dataset show the effectiveness of our proposed Vox-Fusion++ system.

\vspace{1mm}
\noindent\textbf{Surface Reconstruction:} 
In~\autoref{fig:scannet-mesh-comp}, we show the geometric reconstruction results of different scenes using our system. 
It is evident from the comparison that our approach consistently outperforms three baselines, demonstrating its superiority, particularly in reconstructing fine details. 
The effectiveness of our method can be attributed to the expressive power of the signed distance representation and the voxel-based point sampling. 
By leveraging these capabilities, our system is able to reconstruct surfaces with finer details, resulting in more accurate and visually appealing representations.

\begin{figure}
    \centering
    \includegraphics[width=0.9\linewidth]{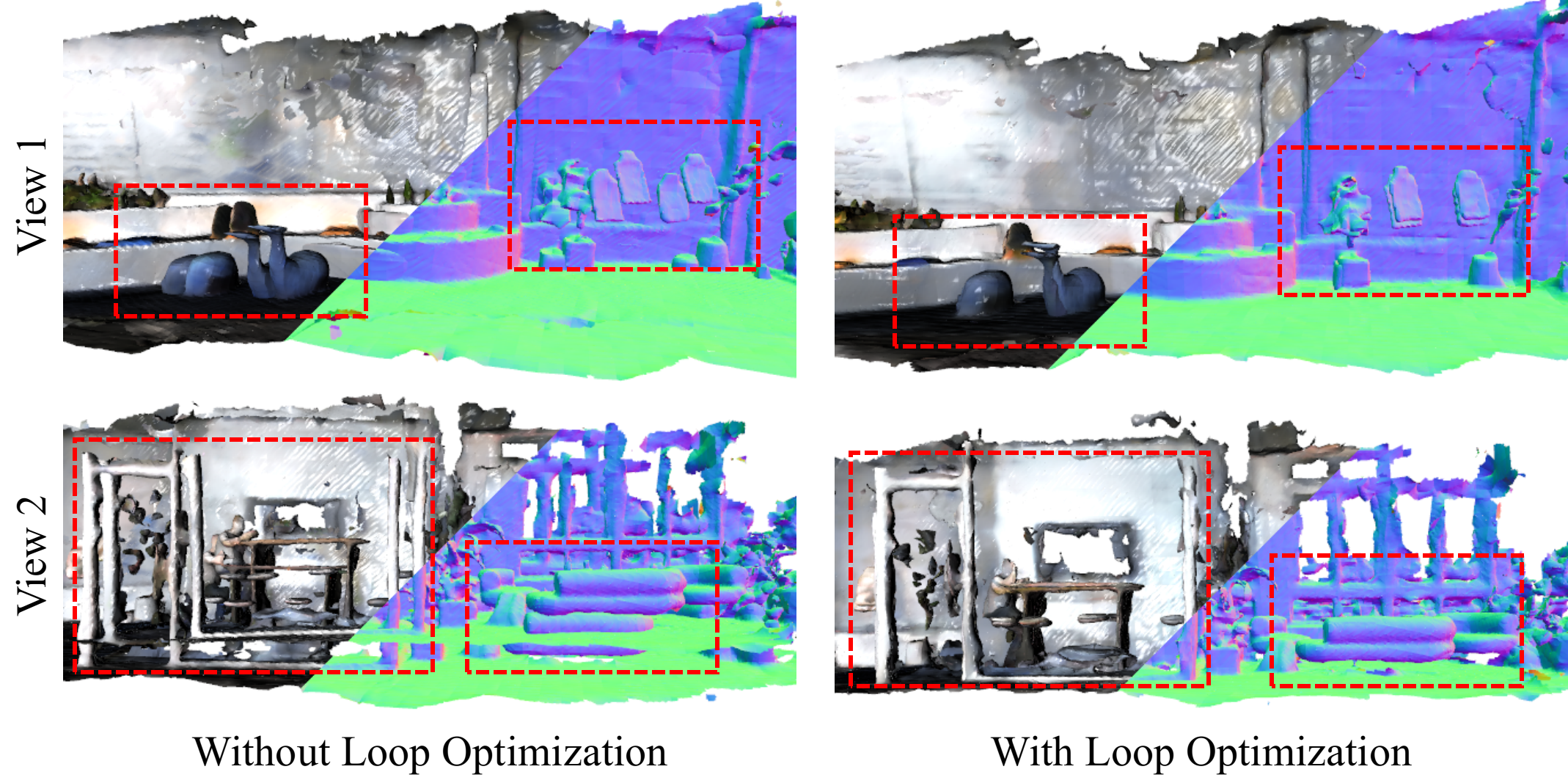}
    \caption{Reconstruction comparison between whether using loop optimization. The red dotted boxes show the difference in details.}
    \label{fig:loop-mesh-diff}
    \vspace{7mm}
\end{figure}

\begin{figure*}
    \centering
    \includegraphics[width=0.8\linewidth]{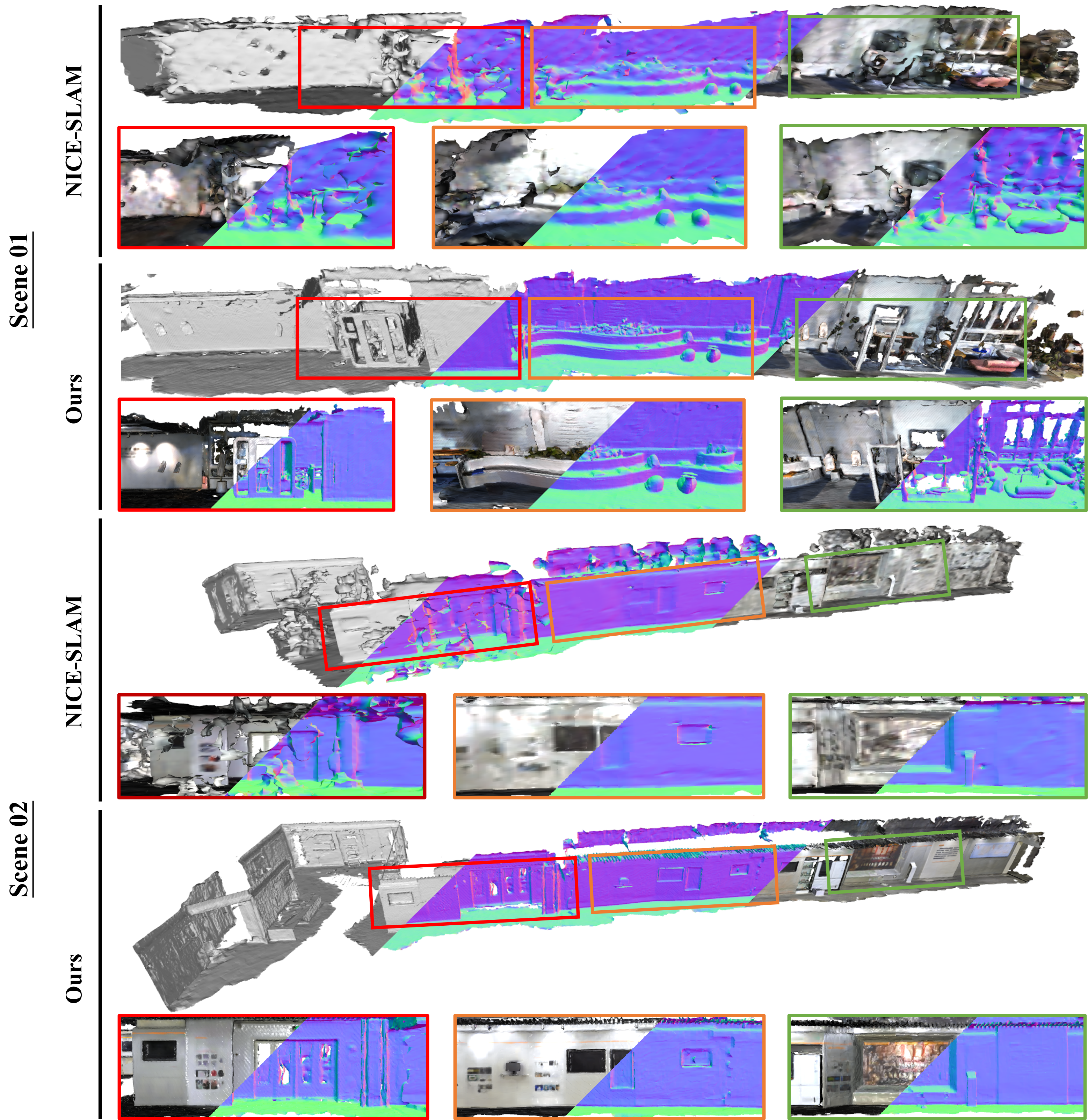}
    \caption{Qualitative comparison on our self-captured \textit{scene01} and \textit{scene02} of NICE-SLAM and Ours. We show the reconstructed mesh of different shading modes from different views. Our approach can achieve much better reconstruction results.}
    \label{fig:mmw-mesh-compar}
\end{figure*}

\subsection{Results on Our Large Indoor Scenes}
To verify the reconstruction results on large indoor scenes, we evaluate our system on two self-captured large indoor sequences with weak textures and weak geometric structures. 
The two scenes are captured from generic office buildings with areas of 35m$\times$5m, and 45m$\times$10m, respectively. 
To validate the effectiveness of our loop detection and optimization method, \textit{scene 01} is deliberately captured moving back-and-forth to include many loop closures.

\vspace{1mm}
\noindent\textbf{Effect of Loop Optimization:} 
~\autoref{fig:loop-mesh-diff} shows the color mesh and normal of the reconstruction without/with our loop optimization from two different views in \textit{scene 01}.
When the same object/scene structures are observed at different moments with accumulated pose drift, at best a duplicate of the object/scene structures tends to appear if no loop optimization is performed, at worst tracking failure will happen.
As can be seen from the figure, performing loop optimizations can reduce pose drift and merge duplicate objects from different views with long intervals. 
{We also show the absolute trajectory error (ATE) in~\autoref{tab:loop_optimization}. The ground truth poses of two scenes are obtained with HLoc~\cite{hloc} (SuperPoint~\cite{superpoint} + SuperGlue~\cite{sarlin2020superglue}).
As can be seen, compared to the original Vox-Fusion~\cite{vox-fusion}, adding loop optimization can lead to trajectory estimation performance improvement.

\vspace{1mm}
\noindent\textbf{Surface Reconstruction:} In addition, we also show the qualitative comparison results of the surface reconstruction.
In \autoref{fig:mmw-mesh-compar}, we show the surface reconstruction results on our captured \textit{scene 01} and \textit{scene 02} of NICE-SLAM and our approach. Note that the tracking process of Co-SLAM failed on two scenes due to the weak texture and fast camera movement.
The local meshes of different views are indicated by different colored boxes. As can be seen from the figure, our approach can recover more accurate color information and smoother surfaces.
Besides, due to the memory requirement of multi-level dense grid representation, NICE-SLAM is unable to reconstruct the last segment of the sequence of \textit{scene 02}. But our incremental mapping strategy not only can reconstruct large scenes but also can produce good color and geometric accuracy.


\begin{figure*}[tbp]
  \centering
  \scriptsize
  \setlength{\tabcolsep}{0.5pt}
  \newcommand{\sz}{0.22}  %
  
  \begin{tabular}{c@{\hspace{3pt}}c@{\hspace{3pt}}c@{\hspace{3pt}}c}
  \includegraphics[width=\sz\textwidth]{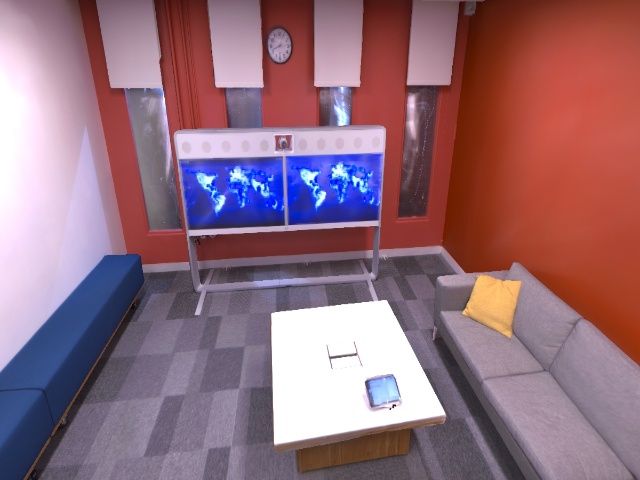} &
  \includegraphics[width=\sz\textwidth]{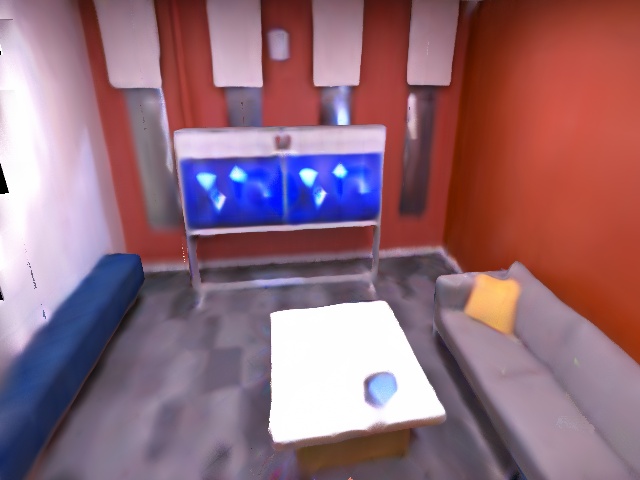} &
  \includegraphics[width=\sz\textwidth]{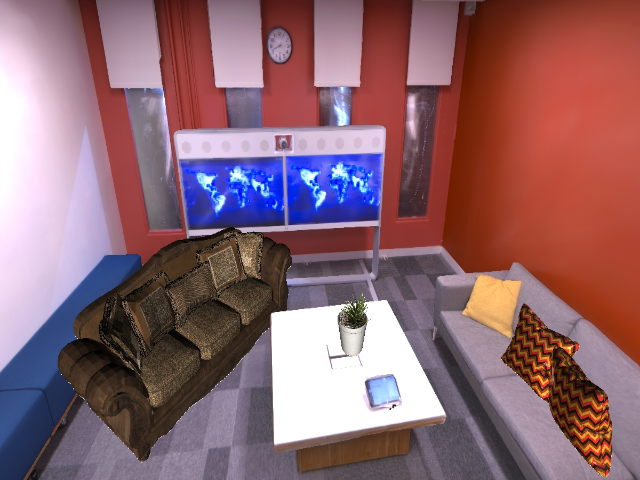} &
  \includegraphics[width=\sz\textwidth]{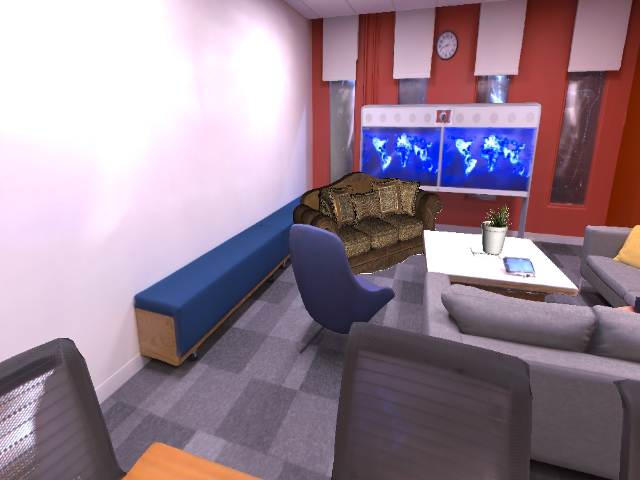} \\
  {\tt \footnotesize{(a) Ground Truth}} & {\tt \footnotesize{(b) Rendered Image}} & {\tt \footnotesize{(c) AR View1}} & {\tt \footnotesize{(d) AR View2}} \\
  \end{tabular}
  \caption{The ground truth image (a) in the Replica office3 dataset and its rendered image (b) with our reconstructed scene. And we place some pre-defined objects in office3, which is shown by (c) and (d) in different viewpoints. We show that we can achieve a good occlusion relationship between real and virtual objects.}
\label{fig:ar_images}
\end{figure*}
\begin{figure*}[!htp]
\centering
\includegraphics[width=0.9\textwidth]{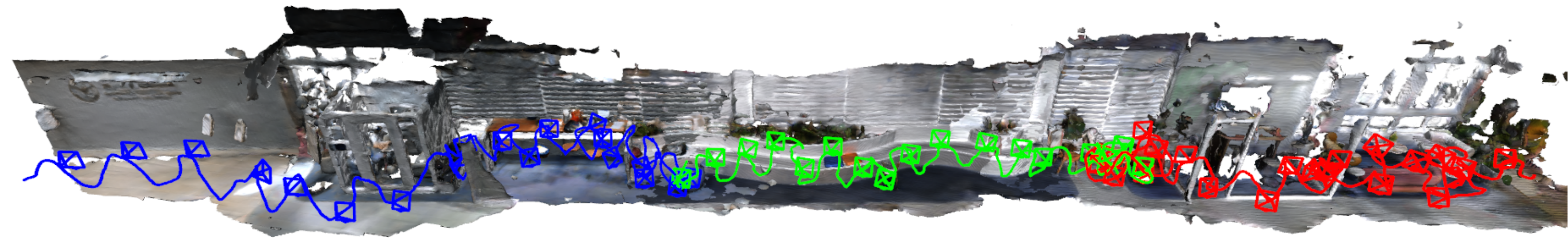}        
\caption{Reconstruction results of collaborative mapping. The whole scene is captured with three different robots (trajectories of different robots are shown in different colors). Our approach can perform collaborative mapping with the ability of loop detection and pose optimization.}
\label{fig:multi-recon}
\end{figure*}

\begin{table}
\caption{Average time spent on each component.}
\label{tab:time}
  \scriptsize%
	\centering%
  \begin{tabular}{%
	l%
	*{7}{c}%
	}
  \toprule
  Components & Measured time \\
  \midrule
    Tracking & 12 ms \\
    Mapping & 55 ms \\
  \midrule
    Voxel allocation & 0.1 ms \\
    Ray-voxel intersection test & 0.9 ms \\
    Point sampling & 1 ms \\
    Volume rendering & 4 ms \\
    Back-propagation & 6 ms \\
  \bottomrule
  \end{tabular}%
\vspace{-1em}
\end{table}

\begin{table}[t]
\caption{Memory consumption for implicit features.}
\label{tab:param}
  \scriptsize%
	\centering%
  \begin{tabular}{%
	l%
	*{2}{c}%
	*{2}{c}%
	*{2}{c}%
	}
  \toprule
  Method & Decoder & Embedding & Total\\
  \midrule
    NICE-SLAM~\cite{zhu:2021:niceslam} & 0.22MB & 238.88MB & 239.10MB \\
    Co-SLAM~\cite{co-slam} & 0.26MB & 1.57MB & 1.83MB \\
    Ours & 0.20MB & 0.15MB & 0.35MB \\
  \bottomrule
  \end{tabular}%
\vspace{-1em}
\end{table}

\subsection{Time and Memory Efficiency}
Our system is designed with a highly efficient multi-process implementation for parallel tracking and mapping. By creating local copies of shared resources like voxels, features, and the implicit decoder during map updates, we minimize resource contention and improve overall performance. In terms of performance evaluation, we have conducted running time analysis and memory comparisons below. 

\vspace{1mm}
\noindent\textbf{Time Analysis:} To analyze the running time of our sparse voxel-based sampling and rendering method, we conducted profiling experiments on the synthetic Replica dataset. We measured the average time spent on crucial components, such as voxel allocation, ray-voxel intersection tests, and volume rendering. The experiments were performed on a single NVIDIA RTX 3090 graphics card. The results, shown in~\autoref{tab:time}, demonstrate that our voxel manipulation functions have minimal impact on the overall running time of the reconstruction pipeline. Depending on the complexity of the scene, our method takes approximately $150$-$200$ ms for tracking a new frame and $450$-$550$ ms for the joint frame and map optimization. In a typical setting, our system achieves a tracking rate of $\sim 5hz$ and an optimization rate of $\sim 2hz$.

\vspace{1mm}
\noindent\textbf{Memory Comparison:} As previously explained, our sparse voxel structure enables us to allocate voxels only for objects and surfaces, which typically represent only a fraction of the entire environment. To compare the memory consumption of our system with NICE-SLAM and Co-SLAM, we profiled the memory usage of implicit decoders and voxel embeddings on the Replica office-0 scene. As shown in~\autoref{tab:param}, it is evident that our method achieves better reconstruction accuracy while utilizing significantly less memory compared to NICE-SLAM and Co-SLAM. NICE-SLAM utilizes four layers of densely populated voxel grids, whereas we only use one layer, which further contributes to our memory-efficient approach. Compared to the hash table used in the Co-SLAM, our system still use fewer voxel embedding features.

\section{Applications}
\label{sec:app}
Our Vox-Fusion++ system excels in accurately estimating camera poses, surface reconstruction, and realistic image rendering. 
Additionally, it enables collaborative mapping with multiple robots through loop detection. 
We present AR and collaborative mapping demonstrations to showcase its capabilities in various real-world scenarios and applications.

\vspace{1mm}
\noindent\textbf{Augmented Reality:} For AR applications, we can place arbitrary virtual objects into real reconstructed scenes, and accurately represent the occlusion relationship between real and virtual contents using the rendered depth maps. 
We show examples of rendered images and AR demo images in~\autoref{fig:ar_images}. 
As can be seen, our dense scene representation allows us to handle occlusion between different objects very well in the AR demo.

\vspace{1mm}
\noindent\textbf{Collaborative Mapping:} Our multi-map incremental mapping approach can also support collaborative mapping with several mapping robots.
When we detect a loop closure between different maps, we can register corresponding maps together using the procedure described in~\autoref{sec:multi-map}. Except that, instead of transforming loop candidate frame pose $T^{c}_{src}$ to the target map pose $T^{m}_{trg}$ given known pose priors, we treat two matching key-frames with the same poses, and propagate pose changes to the maps they associated to, since each agent has a different coordinate system. In practice we found this method to work well because our hierarchical optimization scheme can robustly optimize key-frame poses. We show the reconstruction results of collaborative mapping in ~\autoref{fig:multi-recon}. As shown in the figure, we represent the trajectories of different cameras or robots in different colors. We can register different trajectories together when we find the loop closure or consensus regions.

\section{Conclusion}
\label{sec:limit}

We propose Vox-Fusion++, a novel dense tracking and mapping system built on voxel-based implicit surface representation with multi-maps. 
Our system supports dynamic voxel creation, which is more suitable for practical scenes, we also design a multi-process architecture and corresponding strategies for better performance. 
Besides, we propose loop detection and optimization for large scene reconstruction along with our incremental multi-map strategy. 
Experiments show that our method achieves higher accuracy while using smaller memory and faster speed, we can also correct long-term drifts and achieve collaborative mapping using our proposed loop detection and hierarchical optimization method. 
Currently, our method cannot robustly handle dynamic objects.
We consider these as potential future works.


\bibliographystyle{IEEEtran}
\bibliography{paper}

\begin{thebibliography}{10}
\providecommand{\url}[1]{#1}
\csname url@samestyle\endcsname
\providecommand{\newblock}{\relax}
\providecommand{\bibinfo}[2]{#2}
\providecommand{\BIBentrySTDinterwordspacing}{\spaceskip=0pt\relax}
\providecommand{\BIBentryALTinterwordstretchfactor}{4}
\providecommand{\BIBentryALTinterwordspacing}{\spaceskip=\fontdimen2\font plus
\BIBentryALTinterwordstretchfactor\fontdimen3\font minus
  \fontdimen4\font\relax}
\providecommand{\BIBforeignlanguage}[2]{{%
\expandafter\ifx\csname l@#1\endcsname\relax
\typeout{** WARNING: IEEEtran.bst: No hyphenation pattern has been}%
\typeout{** loaded for the language `#1'. Using the pattern for}%
\typeout{** the default language instead.}%
\else
\language=\csname l@#1\endcsname
\fi
#2}}
\providecommand{\BIBdecl}{\relax}
\BIBdecl

\bibitem{indoor_traj}
Y.~Dai, C.~Wen, H.~Wu, Y.~Guo, L.~Chen, and C.~Wang, ``Indoor 3d human
  trajectory reconstruction using surveillance camera videos and point
  clouds,'' \emph{{IEEE} Trans. Circuits Syst. Video Technol.}, vol.~32, no.~4,
  pp. 2482--2495, 2022.

\bibitem{hand-recon}
H.~Sun, X.~Zheng, P.~Ren, J.~Wang, Q.~Qi, and J.~Liao, ``{SMR:} spatial-guided
  model-based regression for 3d hand pose and mesh reconstruction,''
  \emph{{IEEE} Trans. Circuits Syst. Video Technol.}, vol.~34, no.~1, pp.
  299--314, 2024.

\bibitem{newcombe:2011:dtam}
R.~A. Newcombe, S.~J. Lovegrove, and A.~J. Davison, ``{DTAM: Dense tracking and
  mapping in real-time},'' in \emph{{IEEE} International Conference on Computer
  Vision}, 2011, pp. 2320--2327.

\bibitem{weerasekera:2019:gooddense}
C.~S. Weerasekera, R.~Garg, Y.~Latif, and I.~Reid, ``{Learning Deeply
  Supervised Good Features to Match for Dense Monocular Reconstruction},''
  \emph{Lecture Notes in Computer Science}, vol. 11365 LNCS, pp. 609--624,
  2019.

\bibitem{stuckler:2014:multisurfel}
J.~St{\"{u}}ckler and S.~Behnke, ``{Multi-resolution surfel maps for efficient
  dense {3D} modeling and tracking},'' \emph{Journal of Visual Communication
  and Image Representation}, vol.~25, pp. 137--147, 2014.

\bibitem{whelan:2015:efusion}
T.~Whelan, S.~Leutenegger, R.~F. Salas-Moreno, B.~Glocker, and A.~J. Davison,
  ``{ElasticFusion: Dense SLAM without a pose graph},'' in \emph{Robotics:
  Science and Systems}, vol.~11, 2015.

\bibitem{wang:2019:densesurfel}
K.~Wang, F.~Gao, and S.~Shen, ``{Real-time scalable dense surfel mapping},'' in
  \emph{{IEEE} International Conference on Robotics and Automation}, 2019, pp.
  6919--6925.

\bibitem{newcombe:2011:kinfu}
R.~A. Newcombe, S.~Izadi, O.~Hilliges, D.~Molyneaux, D.~Kim, A.~J. Davison,
  P.~Kohli, J.~Shotton, S.~Hodges, and A.~Fitzgibbon, ``{KinectFusion:}
  real-time dense surface mapping and tracking,'' in \emph{{IEEE} International
  Symposium on Mixed and Augmented Reality}, 2011, pp. 127--136.

\bibitem{niebner:2013:voxelhashing}
M.~Nie\ss{}ner, M.~Zollh\"{o}fer, S.~Izadi, and M.~Stamminger, ``Real-time {3D}
  reconstruction at scale using voxel hashing,'' \emph{ACM Transactions on
  Graphics}, vol.~32, no.~6, 2013.

\bibitem{kahler:2015:inftam}
O.~K{\"{a}}hler, V.~A. Prisacariu, C.~Y. Ren, X.~Sun, P.~H.~S. Torr, and D.~W.
  Murray, ``Very high frame rate volumetric integration of depth images on
  mobile devices,'' \emph{{IEEE} Trans. Vis. Comput. Graph.}, vol.~21, no.~11,
  pp. 1241--1250, 2015.

\bibitem{yang:2022:fdslam}
X.~Yang, Y.~Ming, Z.~Cui, and A.~Calway, ``{FD-SLAM:} 3-{D} reconstruction
  using features and dense matching,'' in \emph{International Conference on
  Robotics and Automation}, 2022, pp. 8040--8046.

\bibitem{high-performance}
G.~Michailidis, R.~Pajarola, and I.~Andreadis, ``High performance stereo system
  for dense 3-d reconstruction,'' \emph{{IEEE} Trans. Circuits Syst. Video
  Technol.}, vol.~24, no.~6, pp. 929--941, 2014.

\bibitem{bloesch:2018:codeslam}
M.~Bloesch, J.~Czarnowski, R.~Clark, S.~Leutenegger, and A.~J. Davison,
  ``{CodeSLAM} - learning a compact, optimisable representation for dense
  visual {SLAM},'' in \emph{{IEEE} Conference on Computer Vision and Pattern
  Recognition}, 2018, pp. 2560--2568.

\bibitem{czarnowski:2020:deepfactors}
J.~Czarnowski, T.~Laidlow, R.~Clark, and A.~Davison, ``Deep{F}actors: Real-time
  probabilistic dense monocular slam,'' \emph{IEEE Robotics and Automation
  Letters}, vol.~5, pp. 721--728, 2020.

\bibitem{matsuki:2021:codemapping}
M.~Hidenobu, S.~Raluca, C.~Jan, and J.~D. Andrew, ``{CodeMapping}: Real-time
  dense mapping for sparse slam using compact scene representations,''
  \emph{{IEEE} Robotics and Automation Letters}, 2021.

\bibitem{mildenhall:2020:nerf}
B.~Mildenhall, P.~P. Srinivasan, M.~Tancik, J.~T. Barron, R.~Ramamoorthi, and
  R.~Ng, ``Nerf: Representing scenes as neural radiance fields for view
  synthesis,'' in \emph{European Conference on Computer Vision}, 2020.

\bibitem{sucar:2021:imap}
E.~Sucar, S.~Liu, J.~Ortiz, and A.~J. Davison, ``{iMAP}: Implicit mapping and
  positioning in real-time,'' in \emph{{IEEE} International Conference on
  Computer Vision}.\hskip 1em plus 0.5em minus 0.4em\relax {IEEE}, 2021, pp.
  6209--6218.

\bibitem{zhu:2021:niceslam}
Z.~Zhu, S.~Peng, V.~Larsson, W.~Xu, H.~Bao, Z.~Cui, M.~R. Oswald, and
  M.~Pollefeys, ``{NICE-SLAM:} neural implicit scalable encoding for {SLAM},''
  in \emph{{IEEE} Conference on Computer Vision and Pattern Recognition}, 2022,
  pp. 12\,786--12\,796.

\bibitem{co-slam}
H.~Wang, J.~Wang, and L.~Agapito, ``{Co-SLAM:} joint coordinate and sparse
  parametric encodings for neural real-time slam,'' in \emph{{IEEE} Conference
  on Computer Vision and Pattern Recognition}, 2023, pp. 13\,293--13\,302.

\bibitem{vox-fusion}
X.~Yang, H.~Li, H.~Zhai, Y.~Ming, Y.~Liu, and G.~Zhang, ``{Vox-Fusion}: Dense
  tracking and mapping with voxel-based neural implicit representation,'' in
  \emph{IEEE International Symposium on Mixed and Augmented Reality}, 2022, pp.
  499--507.

\bibitem{Vox-Surf}
H.~Li, X.~Yang, H.~Zhai, Y.~Liu, H.~Bao, and G.~Zhang, ``{Vox-Surf}:
  Voxel-based implicit surface representation,'' \emph{IEEE Transactions on
  Visualization and Computer Graphics}, pp. 1--12, 2022.

\bibitem{liu:2020:nsvf}
L.~Liu, J.~Gu, K.~Z. Lin, T.~Chua, and C.~Theobalt, ``Neural sparse voxel
  fields,'' in \emph{Annual Conference on Neural Information Processing
  Systems}, 2020.

\bibitem{roth:2012:mkinfu}
H.~Roth and M.~Vona, ``Moving volume kinectfusion,'' in \emph{British Machine
  Vision Conference}, 2012, pp. 1--11.

\bibitem{kerl:2013:dvo}
C.~Kerl, J.~Sturm, and D.~Cremers, ``Dense visual {SLAM} for {RGB-D} cameras,''
  in \emph{{IEEE/RSJ} International Conference on Intelligent Robots and
  Systems}, 2013, pp. 2100--2106.

\bibitem{kahler:2016:inftam2}
O.~K{\"{a}}hler, V.~A. Prisacariu, and D.~W. Murray, ``Real-time large-scale
  dense {3D} reconstruction with loop closure,'' in \emph{European Conference
  on Computer Vision}, vol. 9912, 2016, pp. 500--516.

\bibitem{schops:2019:badslam}
T.~Sch{\"{o}}ps, T.~Sattler, and M.~Pollefeys, ``{BAD} {SLAM:} bundle adjusted
  direct {RGB-D} {SLAM},'' in \emph{{IEEE} Conference on Computer Vision and
  Pattern Recognition}, 2019, pp. 134--144.

\bibitem{li:2015:rgbdreloc}
S.~Li and A.~Calway, ``{RGBD} relocalisation using pairwise geometry and
  concise key point sets,'' in \emph{{IEEE} International Conference on
  Robotics and Automation}.\hskip 1em plus 0.5em minus 0.4em\relax {IEEE},
  2015, pp. 6374--6379.

\bibitem{dai:2017:bundlefusion}
A.~Dai, M.~Nie{\ss}ner, M.~Zollh{\"{o}}fer, S.~Izadi, and C.~Theobalt,
  ``Bundle{F}usion: Real-time globally consistent {3D} reconstruction using
  on-the-fly surface reintegration,'' \emph{{ACM} Trans. Graph.}, vol.~36,
  no.~3, pp. 24:1--24:18, 2017.

\bibitem{Scene_complete}
X.~Gao, S.~Shen, L.~Zhu, T.~Shi, Z.~Wang, and Z.~Hu, ``Complete scene
  reconstruction by merging images and laser scans,'' \emph{{IEEE} Trans.
  Circuits Syst. Video Technol.}, vol.~30, no.~10, pp. 3688--3701, 2020.

\bibitem{mo-slam}
X.~Shao, L.~Zhang, T.~Zhang, Y.~Shen, and Y.~Zhou, ``Mofis-slam: {A}
  multi-object semantic {SLAM} system with front-view, inertial, and
  surround-view sensors for indoor parking,'' \emph{{IEEE} Trans. Circuits
  Syst. Video Technol.}, vol.~32, no.~7, pp. 4788--4803, 2022.

\bibitem{sf-depth}
C.~Tseng and S.~Wang, ``Shape-from-focus depth reconstruction with a spatial
  consistency model,'' \emph{{IEEE} Trans. Circuits Syst. Video Technol.},
  vol.~24, no.~12, pp. 2063--2076, 2014.

\bibitem{huang:2021:difusion}
J.~Huang, S.~Huang, H.~Song, and S.~Hu, ``{DI}-{F}usion: Online implicit 3d
  reconstruction with deep priors,'' in \emph{{IEEE} Conference on Computer
  Vision and Pattern Recognition}, 2021, pp. 8932--8941.

\bibitem{teed2022deep}
Z.~Teed, L.~Lipson, and J.~Deng, ``Deep patch visual odometry,'' \emph{arXiv
  preprint arXiv:2208.04726}, 2022.

\bibitem{teed:2021:droid}
Z.~Teed and J.~Deng, ``{DROID-SLAM:} deep visual {SLAM} for monocular, stereo,
  and {RGB-D} cameras,'' in \emph{Annual Conference on Neural Information
  Processing Systems}, 2021, pp. 16\,558--16\,569.

\bibitem{nicer-slam}
Z.~Zhu, S.~Peng, V.~Larsson, Z.~Cui, M.~R. Oswald, A.~Geiger, and M.~Pollefeys,
  ``{NICER-SLAM:} neural implicit scene encoding for {RGB} {SLAM},''
  \emph{CoRR}, vol. abs/2302.03594, 2023.

\bibitem{nerf-slam}
A.~Rosinol, J.~J. Leonard, and L.~Carlone, ``{NeRF}-{SLAM}: Real-time dense
  monocular {SLAM} with neural radiance fields,'' \emph{CoRR}, vol.
  abs/2210.13641, 2022.

\bibitem{dim-slam}
H.~Li, X.~Gu, W.~Yuan, L.~Yang, Z.~Dong, and P.~Tan, ``Dense {RGB} slam with
  neural implicit maps,'' in \emph{International Conference on Learning
  Representations}, 2023.

\bibitem{Orbeez-slam}
C.~Chung, Y.~Tseng, Y.~Hsu, X.~Q. Shi, Y.~Hua, J.~Yeh, W.~Chen, Y.~Chen, and
  W.~H. Hsu, ``Orbeez-{SLAM}: {A} real-time monocular visual {SLAM} with {ORB}
  features and nerf-realized mapping,'' \emph{CoRR}, vol. abs/2209.13274, 2022.

\bibitem{eslam}
M.~M. Johari, C.~Carta, and F.~Fleuret, ``{ESLAM}: Efficient dense slam system
  based on hybrid representation of signed distance fields,'' in \emph{{IEEE}
  Conference on Computer Vision and Pattern Recognition}, 2023.

\bibitem{orb-slam2}
R.~Mur{-}Artal and J.~D. Tard{\'{o}}s, ``{ORB-SLAM2:} an open-source {SLAM}
  system for monocular, stereo, and {RGB-D} cameras,'' \emph{{IEEE} Trans.
  Robotics}, vol.~33, no.~5, pp. 1255--1262, 2017.

\bibitem{orb-slam3}
C.~Campos, R.~Elvira, J.~J.~G. Rodr{\'{\i}}guez, J.~M.~M. Montiel, and J.~D.
  Tard{\'{o}}s, ``{ORB-SLAM3:} an accurate open-source library for visual,
  visual-inertial, and multimap {SLAM},'' \emph{{IEEE} Trans. Robotics},
  vol.~37, no.~6, pp. 1874--1890, 2021.

\bibitem{go-surf}
J.~Wang, T.~Bleja, and L.~Agapito, ``{GO}-{S}urf: Neural feature grid
  optimization for fast, high-fidelity {RGB-D} surface reconstruction,'' in
  \emph{International Conference on 3D Vision}, 2022, pp. 433--442.

\bibitem{manhattan-sdf}
H.~Guo, S.~Peng, H.~Lin, Q.~Wang, G.~Zhang, H.~Bao, and X.~Zhou, ``Neural 3{D}
  scene reconstruction with the manhattan-world assumption,'' in \emph{{IEEE}
  Conference on Computer Vision and Pattern Recognition}, 2022, pp. 5501--5510.

\bibitem{oechsle:2021:unisurf}
M.~Oechsle, S.~Peng, and A.~Geiger, ``{UNISURF:} unifying neural implicit
  surfaces and radiance fields for multi-view reconstruction,'' in
  \emph{{IEEE/CVF} International Conference on Computer Vision}, 2021, pp.
  5569--5579.

\bibitem{wang:2021:neus}
P.~Wang, L.~Liu, Y.~Liu, C.~Theobalt, T.~Komura, and W.~Wang, ``{NeuS}:
  Learning neural implicit surfaces by volume rendering for multi-view
  reconstruction,'' in \emph{Annual Conference on Neural Information Processing
  Systems 2021}, 2021, pp. 27\,171--27\,183.

\bibitem{azinovic:2022:neuralrgbd}
D.~Azinovi\'c, R.~Martin-Brualla, D.~B. Goldman, M.~Nie{\ss}ner, and J.~Thies,
  ``Neural {RGB-D} surface reconstruction,'' in \emph{{IEEE} Conference on
  Computer Vision and Pattern Recognition}, 2022, pp. 6290--6301.

\bibitem{yu:2022:plenoxels}
{Sara Fridovich-Keil and Alex Yu}, M.~Tancik, Q.~Chen, B.~Recht, and
  A.~Kanazawa, ``Plenoxels: Radiance fields without neural networks,'' in
  \emph{{IEEE} Conference on Computer Vision and Pattern Recognition}, 2022.

\bibitem{yu:2021:plenoctrees}
A.~Yu, R.~Li, M.~Tancik, H.~Li, R.~Ng, and A.~Kanazawa, ``Plen{O}ctrees for
  real-time rendering of neural radiance fields,'' in \emph{{IEEE}
  International Conference on Computer Vision}.\hskip 1em plus 0.5em minus
  0.4em\relax {IEEE}, 2021, pp. 5732--5741.

\bibitem{takikawa:2021:nglod}
T.~Takikawa, J.~Litalien, K.~Yin, K.~Kreis, C.~T. Loop, D.~Nowrouzezahrai,
  A.~Jacobson, M.~McGuire, and S.~Fidler, ``Neural geometric level of detail:
  Real-time rendering with implicit 3d shapes,'' in \emph{{IEEE} Conference on
  Computer Vision and Pattern Recognition}, 2021, pp. 11\,358--11\,367.

\bibitem{h2mapping}
C.~Jiang, H.~Zhang, P.~Liu, Z.~Yu, H.~Cheng, B.~Zhou, and S.~Shen,
  ``H2-{M}apping: Real-time dense mapping using hierarchical hybrid
  representation,'' \emph{arXiv preprint arXiv:2306.03207}, 2023.

\bibitem{imtooth}
H.~Li, H.~Zhai, X.~Yang, Z.~Wu, Y.~Zheng, H.~Wang, J.~Wu, H.~Bao, and G.~Zhang,
  ``Im{T}ooth: Neural implicit tooth for dental augmented reality,''
  \emph{{IEEE} Trans. Vis. Comput. Graph.}, vol.~29, no.~5, pp. 2837--2846,
  2023.

\bibitem{rematas:2022:urf}
K.~Rematas, A.~Liu, P.~P. Srinivasan, J.~T. Barron, A.~Tagliasacchi,
  T.~Funkhouser, and V.~Ferrari, ``Urban radiance fields,'' in \emph{{IEEE}
  Conference on Computer Vision and Pattern Recognition}, 2022.

\bibitem{strasdat:2011:doublewindow}
H.~Strasdat, A.~J. Davison, J.~M. Montiel, and K.~Konolige, ``{Double window
  optimisation for constant time visual SLAM},'' \emph{{IEEE} International
  Conference on Computer Vision}, pp. 2352--2359, 2011.

\bibitem{vespa:2018:supereight}
E.~Vespa, N.~Nikolov, M.~Grimm, L.~Nardi, P.~H.~J. Kelly, and S.~Leutenegger,
  ``Efficient octree-based volumetric {SLAM} supporting signed-distance and
  occupancy mapping,'' \emph{{IEEE} Robotics Autom. Lett.}, vol.~3, no.~2, pp.
  1144--1151, 2018.

\bibitem{netvlad}
R.~Arandjelovic, P.~Gron{\'{a}}t, A.~Torii, T.~Pajdla, and J.~Sivic,
  ``{NetVLAD}: {CNN} architecture for weakly supervised place recognition,'' in
  \emph{{IEEE} Conference on Computer Vision and Pattern Recognition}, 2016,
  pp. 5297--5307.

\bibitem{patch-netvlad}
S.~Hausler, S.~Garg, M.~Xu, M.~Milford, and T.~Fischer, ``{Patch-NetVLAD}:
  Multi-scale fusion of locally-global descriptors for place recognition,'' in
  \emph{{IEEE} Conference on Computer Vision and Pattern Recognition}, 2021,
  pp. 14\,141--14\,152.

\bibitem{hloc}
P.~Sarlin, C.~Cadena, R.~Siegwart, and M.~Dymczyk, ``From coarse to fine:
  Robust hierarchical localization at large scale,'' in \emph{{IEEE} Conference
  on Computer Vision and Pattern Recognition}, 2019, pp. 12\,716--12\,725.

\bibitem{julian:2019:replica}
J.~Straub, T.~Whelan, L.~Ma, Y.~Chen, E.~Wijmans, S.~Green, J.~J. Engel,
  R.~Mur-Artal, C.~Ren, S.~Verma, A.~Clarkson, M.~Yan, B.~Budge, Y.~Yan,
  X.~Pan, J.~Yon, Y.~Zou, K.~Leon, N.~Carter, J.~Briales, T.~Gillingham,
  E.~Mueggler, L.~Pesqueira, M.~Savva, D.~Batra, H.~M. Strasdat, R.~D. Nardi,
  M.~Goesele, S.~Lovegrove, and R.~Newcombe, ``The {R}eplica dataset: A digital
  replica of indoor spaces,'' \emph{arXiv preprint arXiv:1906.05797}, 2019.

\bibitem{dai:2017:scannet}
A.~Dai, A.~X. Chang, M.~Savva, M.~Halber, T.~Funkhouser, and M.~Nießner,
  ``Scan{N}et: Richly-annotated {3D} reconstructions of indoor scenes,'' in
  \emph{{IEEE} Conference on Computer Vision and Pattern Recognition}, 2017,
  pp. 2432--2443.

\bibitem{sturm:2012:tumrgbd}
J.~Sturm, N.~Engelhard, F.~Endres, W.~Burgard, and D.~Cremers, ``A benchmark
  for the evaluation of {RGB-D} {SLAM} systems,'' in \emph{{IEEE/RSJ}
  International Conference on Intelligent Robots and Systems}, 2012, pp.
  573--580.

\bibitem{superpoint}
D.~DeTone, T.~Malisiewicz, and A.~Rabinovich, ``{SuperPoint}: Self-supervised
  interest point detection and description,'' in \emph{{IEEE} Conference on
  Computer Vision and Pattern Recognition Workshops}, 2018, pp. 224--236.

\bibitem{sarlin2020superglue}
P.-E. Sarlin, D.~DeTone, T.~Malisiewicz, and A.~Rabinovich, ``{SuperGlue}:
  Learning feature matching with graph neural networks,'' in \emph{{IEEE}
  Conference on Computer Vision and Pattern Recognition}, 2020.

\end{thebibliography}

\end{document}